\documentclass{article}

\usepackage{time_2024}
\usepackage{graphicx}
\usepackage{setspace}
\usepackage{multirow}
\usepackage{amsmath}
\usepackage{amsthm}
\usepackage{algorithm}
\usepackage{algpseudocode}
\usepackage{booktabs}
\usepackage{mathtools,amssymb}
\usepackage{subfigure}
\usepackage{booktabs} 
\usepackage{bm,bbm}
\usepackage{multirow}
\usepackage{wrapfig}
\usepackage{hyperref}       
\usepackage{url}            
\usepackage{threeparttable}
\algrenewcommand\algorithmicindent{0.5em}%

\newcommand{\modelname}{TimeDiT}
\newcommand{\modelnamespace}{TimeDiT }

\usepackage[utf8]{inputenc} 
\usepackage[T1]{fontenc}    
\usepackage{hyperref}       
\usepackage{url}            
\usepackage{booktabs}       
\usepackage{amsfonts}       
\usepackage{nicefrac}       
\usepackage{microtype}      
\usepackage{xcolor}         
\usepackage{algpseudocode}
\usepackage{algorithm}
\usepackage{natbib}

\newtheorem{theorem}{Theorem}[section]

\def\modelnamemor{\textsc{Moirai}}
\def\smallmodel{{\modelnamemor}\textsubscript{Small}}
\def\basemodel{{\modelnamemor}\textsubscript{Base}}
\def\largemodel{{\modelnamemor}\textsubscript{Large}}

\usepackage{titletoc}
\usepackage{appendix}
\usepackage{etoc}

\title{TimeDiT: General-purpose Diffusion Transformers for Time Series Foundation Model}

\author{
  Defu Cao\thanks{Equal Contribution with Alphabetical Order.}, Wen Ye\footnotemark[1], Yizhou Zhang, Yan Liu \\
  Department of Computer Science\\
  University of Southern California\\
  Los Angeles, CA 90007 \\
  \texttt{\{defucao, yewen, zhangyiz, yanliu.cs\}@usc.edu} \\
}

\begin{document}

\maketitle

\begin{abstract}

Foundation models, particularly Large Language Models (LLMs), have revolutionized text and video processing, yet time series data presents distinct challenges for such approaches due to domain-specific features such as missing values, multi-resolution characteristics, etc.
Furthermore, the de-facto autoregressive transformers tend to learn deterministic temporal dependencies within pre-trained data while overlooking inherent uncertainties and lacking integration of physical constraints.
In this paper, we introduce TimeDiT, a diffusion transformer model that synergistically combines transformer-based temporal dependency learning with diffusion-based probabilistic sampling. TimeDiT employs a unified masking mechanism to harmonize the training and inference process across diverse tasks while introducing a theoretically grounded, finetuning-free model editing strategy that enables flexible integration of external knowledge during sampling. 
 Acknowledging the challenges of unifying multiple downstream tasks under a single model, our systematic evaluation demonstrates TimeDiT's effectiveness both in fundamental tasks, i.e., forecasting and imputation, through zero-shot/fine-tuning; and in domain tasks, i.e., multi-resolution forecasting, anomaly detection, and data generation, establishing it as a \textit{proto-foundation model} that bridges the gap between general-purpose and domain-specific models.

\end{abstract}

\section{Introduction}
Time series analysis is fundamental across natural science, sustainability, and healthcare \citep{ye2024beyond, kamra2021polsird, cuomo2022scientific, burgertowards}. While specialized models like TCNs \citep{tcn}, LSTMs \citep{lstm}, GNNs \citep{GNN_1}, and Transformers \citep{zhang2022crossformer} have advanced in time series tasks including forecasting, imputation, anomaly detection, and data generation, etc, their domain-specific design limits broader applicability. Inspired by the success of pre-trained models such as Deepseek~\cite{liu2024deepseek}, GPT-4~\citep{GPT-4}, LLaMA \citep{Touvron2023LLaMAOA} and Vision Transformer \citep{dosovitskiy2020vit}'s success in addressing foundation problems in language and vision, researchers have begun exploring universal time series forecasting models \citep{ansari2024chronos, liu2024timer, gruver2024large}. While these initial efforts have shown promise in temporal dependency forecasting, developing a foundation model that adapts to diverse time series tasks without \textit{task-specific customization} remains challenging.

Recent advances in time series forecasting foundation models have established promising directions, while also revealing opportunities for further enhancement. 
Current approaches employ diverse tokenization strategies—such as TimeMoE~\citep{shi2024timemoe} and TimesFM~\cite{das2023decoder}'s patching, and Moirai~\cite{woo2024unified}'s multiple patching—which effectively manage varying sequence lengths based on fixed patch length, yet present opportunities for improved generalization across dynamically shifting data characteristics. The channel independence strategies \citep{Yuqietal-2023-PatchTST} adopted by models like Timer~\citep{liu2024timer} and Chronos~\cite{ansari2024chronos} enable efficient model scaling, while suggesting potential for enhanced modeling of the intricate interplay between temporal patterns and cross-channel dependencies in real-world time series data (TSD). Moreover, conventional auto-regressive based models typically learn a deterministic, unique mapping relationship from historical data, limiting their ability to capture the inherent uncertainties and stochastic nature of TSD.

Moreover, real-world time series data poses fundamental challenges that disrupt inherent pattern learning in transformer models. \textit{Missing values}~\citep{kollovieh2023predict} break temporal continuity, \textit{multi-resolution sampling}~\citep{niu2023time} distorts cross-variable relationships, and \textit{irregular temporal intervals}~\citep{cao2023estimating} compromise temporal dependency learning. 
However, current benchmark datasets \citep{li2018dcrnn_traffic, zhou2021informer, gluonTS} often fail to reflect such real-world TSD's complexities, potentially leading to models that underperform in practical applications. In addition, time series processes are often governed by underlying \textit{physical principles}~\citep{meng2022physics}. Incorporating physics knowledge can further enhance model performance and interpretability, especially in data-scarce domains. 
 Addressing these challenges demands innovative solutions across data preprocessing, model architecture design, and training strategies to create robust models that can effectively handle diverse TSD with varying temporal and feature characteristics.

While a universal foundation model for all time series tasks remains an aspirational goal, we propose a pragmatic yet powerful solution that bridges foundation models with task-specific adaptability, addressing both immediate challenges and future scalability. The solution is designed to espouse these three foundational \textbf{principles}: First, comprehensive \textit{pre-training} through extensive data exposure establishes a robust foundation for temporal pattern recognition. This foundational capability directly addresses fundamental tasks with uniform definitions and minimal constraints, enabling forecasting and imputation through zero-shot inference or fine-tuning. Second, for tasks requiring \textit{in-domain training}, such as anomaly detection where patterns inherently vary across datasets, adaptable pipelines are needed that maintain consistency without model architecture updating. Third, the foundation model can \textit{incorporate physical constraints} and domain knowledge, enhancing performance while ensuring solutions remain grounded in real-world applicability.

Following the aforementioned foundational principles, in this work, we introduce \modelnamespace and position it as a proto-foundation model designed { to process practical TSD across domains, frequencies, and sampling patterns}. As a diffusion transformer-based~\citep{Peebles2022DiT} approach, TimeDiT combines the transformer architecture's generalizability and expertise in capturing temporal dependencies with diffusion models' capacity to explore diverse solutions within a broad prior space, enabling the direct generation of high-quality samples.
\modelnamespace incorporates comprehensive time series mask units, including position mask, stride mask and block mask, for both task-agnostic pre-training and task-specific inference that offer flexibility in handling varying input shapes and enable self-supervised learning (SSL).
Furthermore, during the sampling stage, TimeDiT can incorporate physics knowledge, such as pre-defined partial differential equations (PDEs), as a theoretically grounded energy-based prior, supported by theoretical guarantees. This approach guides the reverse diffusion process using physics-based constraints, resulting in generated samples that adhere to known physical laws and domain-specific requirements, thereby enhancing sample quality and model applicability across various scientific and engineering contexts. This principled approach naturally leads to our comprehensive three-part experimental evaluation: First, zero-shot and fine-tuning on canonical tasks including forecasting and imputation from electricity, traffic, climate, and finance domains. Second, domain-adaptive tasks require prior knowledge of data characteristics—specifically multi-resolution forecasting, anomaly detection, and targeted data generation—including climate, health care, finance, energy, hydraulics, and science domains. Third, we further examine the model's capacity for knowledge integration through physics-informed constraints and multimodal fusion.  Notably, the results of zero-shot experiments show that our model can be used as a foundation model even without fine-tuning, although fine-tuning may be necessary in some cases. In addition, TimeDiT achieved new state-of-the-art (SOTA) performance in uncertainty quantification (UQ) across real-world datasets for probabilistic forecasting with missing values or multi-resolution. The main contribution of our work can be summarized as the following threefold:
\begin{itemize}
    \item We introduce TimeDiT, a proto-foundation model that integrates transformer-based temporal modeling with diffusion sampling capabilities. Enhanced by unified masking mechanisms, TimeDiT transcends conventional foundation models' focus on forecasting, enabling comprehensive SSL for diverse time series tasks.

    \item We develop a model-editing-free physics knowledge injection framework that leverages physical equations (such as PDEs) as energy-based priors. This approach ensures that generated samples adhere to known physical laws while maintaining model integrity.

    \item We validate TimeDiT through comprehensive experiments aligned with our foundational principles, evaluating its performance across zero-shot, fine-tuning, and in-domain settings. The results demonstrate state-of-the-art performance, cross-task adaptability, and seamless integration of domain knowledge.
  
\end{itemize}

\section{Related Work}
\paragraph{Large Pre-trained Time Series Model}
In the past decades, researchers have excelled in designing sophisticated models for specific time series analysis tasks~\citep{zhang2024multi,fan2024addressing, cao2020spectral, bi2023accurate, zhang2021vigdet,ye2022spatiotemporal, GPT4MTS}. However, the recent emergence of LLMs has inspired the development of general-purpose time series models and the field of time series has seen tremendous exploration efforts towards foundation models~\citep{zerveas2021transformer,zhang2024self} .  Specifically, \citep{gruver2024large} simply encodes time series as strings while TimeLLM \citep{jin2023time} convertes time series into language representations by alignment. TEMPO \citep{cao2023tempo} and S$^2$IP-LLM \citep{pan2024s} further incorporate decomposition technique and prompt design and generalize to unseen data and multimodal scenarios. Additionally, many studies start to follow a two-stage training paradigm of pretraining and finetuning \citep{chang2023llm4ts, dong2024simmtm}. However, previous works including Timer~\citep{liu2024timer}, TimeMoE~\cite{shi2024timemoe}, Chronos~\citep{ansari2024chronos}, TimeGPT~\citep{garza2023timegpt}, UniTime~\citep{liu2024unitime}, TTM~\citep{ekambaram2024ttms} and Moirai~\citep{woounified} mainly focus on the forecasting task only. \citep{zhou2023one} first adapted GPT2 as a general-purpose time series analysis model and extended it to various time series tasks. \citep{talukder2024totem} leveraged VQVAE as a tokenizer for transformer to handle time series tasks and \citep{ansari2024chronos} employed a scaling and quantization technique to embed time series. For more detailed literatures of the general-purpose and foundation time series model, please refer to recent surveys~\citep{liang2024foundation, jin2024position, jiang2024empowering}.

\paragraph{Diffusion models for Time Series}
Despite growing interest in diffusion models across various scenarios~\citep{li2022diffusion, lu2024vdt, sui2024bitsfusion,sui2024disdet}, their application in time series analysis remains less explored compared to pre-trained language models.  Most existing studies also focus solely on forecasting and the choice of backbone model also varies among VAE\citep{li2022generative}, RNN\citep{rasul2021autoregressive}, and transformers. Recently, CSDI \citep{tashiro2021csdi} first utilizes a diffusion model for time series imputation with a self-supervised approach. SSSD~\citep{alcarazdiffusion} combines the structured state space model with the diffusion model for imputation. ImDiffusion~\citep{chen2023imdiffusion} leverages diffusion models as time series imputers to achieve accurate anomaly detection.  D$^3$VAE \citep{li2022generative}  proposes a generative time series forecasting method on top of VAE equipped with the diffusion model. Meanwhile,  DiffusionTS~\citep{yuan2024diffusion} incorporates decomposition into the diffusion model to improve interoperability. 
Although TSDiff~\citep{kollovieh2023predict} build a diffusion pipeline for multiple tasks with refinement, they still train different models for each task. Based on our knowledge, no unified diffusion transformer model has yet been explored for a comprehensive set of time series tasks. For a thorough literature review on diffusion models in time series analysis, please refer to ~\citep{yang2024survey}.

\section{Methodology}
\begin{figure*}
    \centering
    \includegraphics[width = 0.9\linewidth]{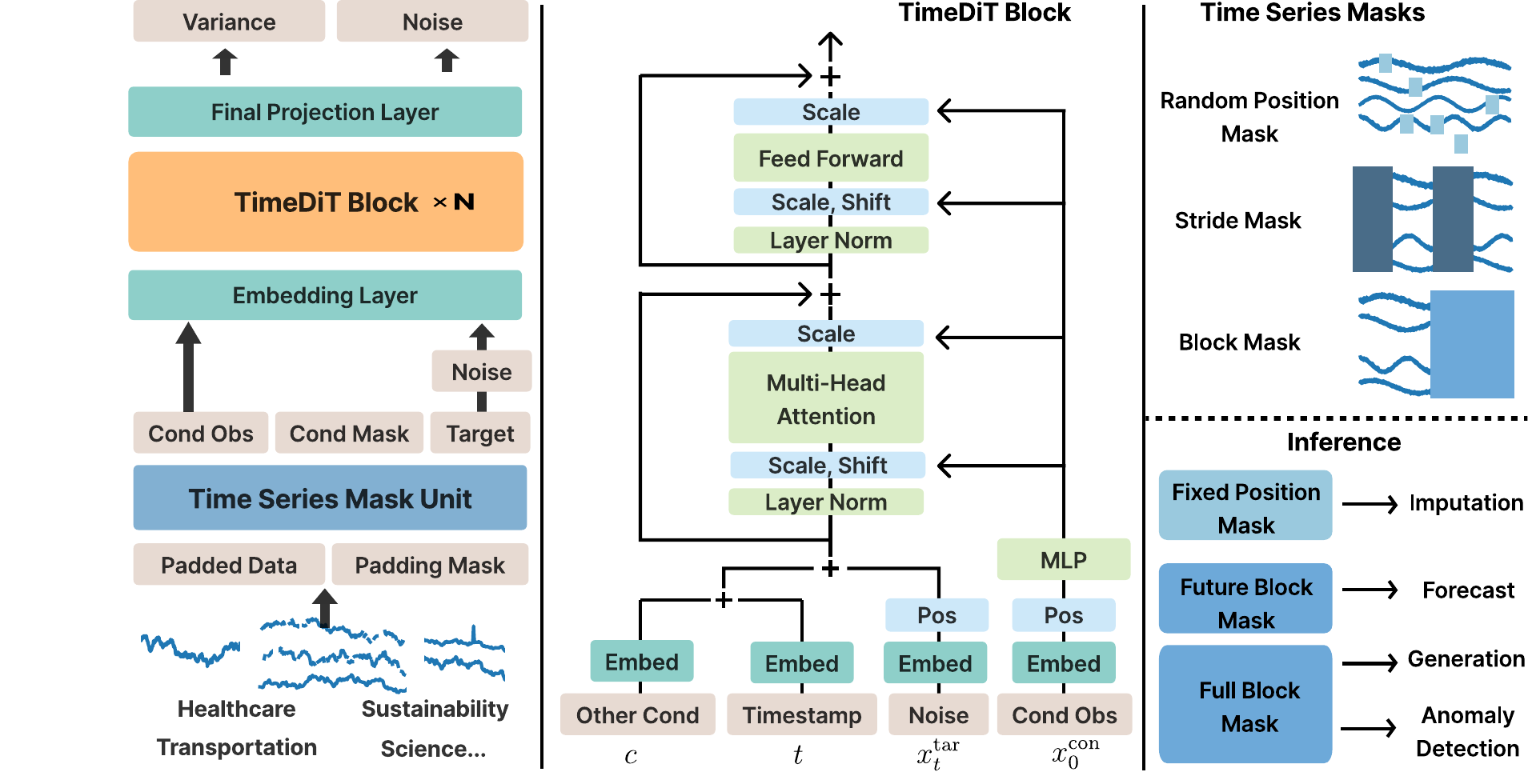}
    \caption{\modelnamespace  Architecture. \underline{Left}: \modelnamespace framework with diverse multivariate time series from different domains with multi-resolution or missing values; \underline{Middle}: Structure of \modelnamespace block; \underline{Right} top: Illustration of masks generated by Time Series Mask Unit; 
    \underline{Right} bottom: Masks for downstream tasks that \modelnamespace handles during inference.}
    \label{fig:model_overview}
\end{figure*}
In this section, we present our main contributions: the proposed foundation model, \modelnamespace, a diffusion model with the transformer backbone designed for multiple time series tasks.
We first outline the uniform problem setting for multiple downstream tasks and offer an in-depth examination of the model architecture. Subsequently, we delve into the training pipeline with mask strategies, which help to build the training scheme in self-supervised learning for time series. Next, we present how to incorporate external information to improve the model's performance during inference stages by generating samples that better conform to real-world requirements. These extensions showcase the flexibility and adaptability of our proposed model, making it a powerful foundation model for a wide range of time series applications.

\subsection{Problem Definition}

We denote a multivariate time series as \( \mathbf{X} = \{x_{i,j}\} \in \mathbb{R}^{K \times L} \), where \( K \) is the number of features and \( L \) is the length of the time series. Each individual entry \( x_{i,j} \) represents the \( j \)-th feature at time step \( i \), for \( i \in \{1, \dots, L\} \) and \( j \in \{1, \dots, K\} \). 
We define an observation mask \( \mathbf{M_{obs}} = \{m_{i,j}\} \in \{0, 1\}^{K \times L} \), where \( m_{i,j} = 0 \) if \( x_{i,j} \) is missing, otherwise, \( m_{i,j} = 1 \).
Let  \( \mathbf{x}_0^{\text{obs}} \in X^{\text{obs}} \) denote the observed subsequence; \( \mathbf{x}_0^{\text{tar}} \) denote the target subsequence of $\mathbf{x}_0^{\text{obs}}$ which could be forecast target or imputation target or the whole sequence depending on the task. Let  \( \mathbf{x}_0^{\text{con}} \) denote the unmasked partial observations in $\mathbf{x}_0^{\text{obs}}$ which acts like self-conditions for the masked area  $\mathbf{x}_0^{\text{tar}}$.  Let us use all subscripts of $\mathbf{x}$ to denote diffusion timestamp, and a subscript of 0 means no noise has been applied to the original data. Formally, the goal of our task is to approximate the true conditional time series distribution given the conditional information
$q_{\mathbf{X}}\left(\mathbf{x}_0^{\mathrm{tar}}\mid\mathbf{x}_0^{\mathrm{con}}\right)$
with a model distribution \( p_{\theta}(\mathbf{x}^{\text{tar}}_0 \mid \mathbf{x}^{\text{con}}_0) \), which can be calculated by a diffusion model with conditional information:
\begin{align}
\textstyle    p_\theta\left(\mathbf{x}_{0: T}^{\mathrm{tar}} \mid \mathbf{x}_0^{\mathrm{con}}\right) := p\left(\mathbf{x}_T^{\mathrm{tar}}\right) \prod_{t=1}^T p_\theta\left(\mathbf{x}_{t-1}^{\mathrm{tar}} \mid \mathbf{x}_t^{\mathrm{tar}}, \mathbf{x}_0^{\mathrm{con}}\right), \mathbf{x}_T^{\mathrm{tar}} \sim \mathcal{N}(\mathbf{0}, \mathbf{I}), \text{where} \nonumber\\
p_\theta\left(\mathbf{x}_{t-1}^{\mathrm{tar}} \mid \mathbf{x}_t^{\mathrm{tar}}, \mathbf{x}_0^{\mathrm{con}}\right) := \mathcal{N}\left(\mathbf{x}_{t-1}^{\mathrm{tar}} ; \boldsymbol{\mu}_\theta\left(\mathbf{x}_t^{\mathrm{tar}}, t \mid \mathbf{x}_0^{\mathrm{con}}\right), \sigma_\theta\left(\mathbf{x}_t^{\mathrm{tar}}, t \mid \mathbf{x}_0^{\mathrm{con}}\right) \mathbf{I}\right).
\label{cond_fwd}
\end{align}
The mask mechanism $\mathbf{M}$ plays a critical role in identifying the positions of $\mathbf{x}_0^{\mathrm{con}}$ and $\mathbf{x}_0^{\mathrm{tar}}$. By leveraging these positional differences, our model can adeptly adapt to tasks like forecasting, imputation, and anomaly detection in a unified framework.

\subsection{Time Series Diffusion Transformer}
\label{sec_diff_trans}
Figure \ref{fig:model_overview} shows the overall framework of \modelname. Firstly, we establish $\mathbf{M_{obs}}$ and $\mathbf{x}_0^{\text{obs}}$ based on inputs with varying shapes, missing values, and multi-resolution data. By injecting placeholders, we identify corresponding positions and standardize input shapes across different time series, enabling more efficient and consistent processing.
Then, the unified time series mask unit constructs $\mathbf{M}$ and adapts to diverse scenarios, generating $\mathbf{x}_0^{\text{con}}$ and $\mathbf{x}_0^{\text{tar}}$ with shape $\mathbb{R}^{B \times L \times K}$, where $B$ is the batch size. This enables TimeDiT to learn robust representations in a self-supervised manner by reconstructing the original sequence through denoising $\mathbf{x}_T^{\text{tar}}$. {Adopting a "What You See Is What You Get" (WYSIWYG) design philosophy, our model represents tokens as direct, contiguous arrays of inputs.} After that, the embedding layer {with linear projection maps } $\mathbf{x}_0^{\text{con}}$ and the noised $\mathbf{x}_0^{\text{tar}}$ into a continuous token space without vector quantization~\citep{li2024autoregressive}, thereby preserving input integrity. To model the per-token probability distribution, 
 the \modelnamespace block is designed to autonomously learn cross-channel and temporal correlations through end-to-end training.


\textbf{Diffusion process.} TimeDiT unconditional diffusion process comprises a forward process that gradually adds noise to a data sample \(x_0 \sim q(x)\), transforming it into Gaussian noise \(x_T \sim \mathcal{N}(0, I)\) as defined by Eq.~\ref{fwd} and a reverse denoising process learned by a neural network (Eq.~\ref{bwd}).  To guide samples toward regions of high classifier likelihood, a self-conditional component $\mathbf{x}^{\text{con}}_0$ is integrated.  We can train the denoising model $\boldsymbol{\mu}_\theta\left(\mathbf{x}_t^{\mathrm{tar}}, \mathbf{x}_0^{\mathrm{con}}\right)$ in Eq. ~\ref{cond_fwd} using a weighted mean squared error (MSE) loss, which can be justified as optimizing a weighted variational lower bound on the data log-likelihood:
\begin{equation}\label{loss}
  \textstyle L(\mathbf{x}^{\text{con}}_0) = \sum_{t=1}^{T} \mathbb{E}_{q(\mathbf{x}^{\text{tar}}_t | \mathbf{x}^{\text{con}}_0)} \|\mu(\mathbf{x}^{\text{tar}}_t, \mathbf{x}^{\text{con}}_0) - \mu_\theta(\mathbf{x}^{\text{tar}}_t, t|\mathbf{x}^{\text{con}}_0)\|^2,
\end{equation}
where $\mu(\mathbf{x}^{\text{tar}}_t, \mathbf{x}^{\text{con}}_0)$ is the mean of the posterior \(q(x^{\text{tar}}_{t-1}|\mathbf{x}^{\text{con}}_0, \mathbf{x}^{\text{tar}}_t)\). 

\textbf{Transformer-based Condition Injection.} TimeDiT employs a transformer-based architecture to process multivariate time series data. 
We feed the embedding of noised target series $\mathbf{x}^{\text{tar}}_t$ (with noise schedule $\beta_t \in (0, 1)$), and conditional observation $\mathbf{x}^{\text{con}}_0$ into the TimeDiT block, where the multi-head attention aims to then learns complex relationships within the data.
During the diffusion process, unlike previous approaches~\citep{Peebles2022DiT, lu2024vdt}, we innovatively inject diffusion time information directly into the target noise as these represent universal information across the noised series.
For self-conditional information, 
 while a straightforward approach would be to include conditional information directly in the input sequence through concatenation~\citep{rombach2022high}, we employ adaptive layer normalization (AdaLN) to control the scale and shift of $x_0^{\text{tar}}$ using partial observations $x_0^{\text{con}}$:
 \begin{equation}
    \text{AdaLN}(h,c) = c_{scale} \text{LayerNorm}(h) +c_{shift},
\end{equation}
where \( h \) is the hidden state and \( c_{\text{scale}} \) and \( c_{\text{shift}} \) are the scale and shift parameters derived from the $x_0^{\text{con}}$.
 This method proved empirically more effective than simple input concatenation, as it leverages the scale and shift of $x_0^{\text{con}}$, which are crucial for capturing temporal continuity and progression.

\textbf{Time Series Mask Unit.}
The Time Series Mask Unit is a key component of our model, designed to enhance its versatility and performance across various time series tasks. This unified mechanism incorporates multiple mask types that seamlessly integrate with the model throughout its lifecycle - from self-supervised task-agnostic pre-training to task-specific fine-tuning and inference. The time series mask unit generates four distinct mask types: random mask $\mathbf{M}^{\text{R}}$, block mask $\mathbf{M}^{\text{B}}$, stride mask $\mathbf{M}^{\text{S}}$, and reconstruction mask $\mathbf{M}^{\text{Rec}}$. 
During task-agnostic pre-training, these masks help the model develop robust and generalizable features from the input data, improving overall time series representation. In task-specific training, the masks adapt to the unique requirements of common downstream tasks such as forecasting and imputation, enabling the model to specialize effectively.

As shown in Figure \ref{fig:model_overview} right top, given $\mathbf{x} \in \mathbb{R}^{K\times L}$, the random mask $\text{M}^{\text{R}}$ can be generated by:
\begin{equation}
    \mathbf{M}^{\text{R}}(x,r) = \begin{cases} 1 & z_{i,j}>r, z\in \mathbb{R}^{K\times L}, z\sim Uniform(0,1) \\ 0, & otherwise, \end{cases}
\end{equation}
where $r$ is the mask ratio. For task-specific training and inference, we allow the user to supply customized imputation masks, which replace the random position masks, that could handle the naturally missing data and multi-resolution cases. 
In addition, block mask $\mathbf{M}^{\text{B}}$ can be generated via:
\begin{equation}
    \mathbf{M}^{\text{B}}(x,l) = \begin{cases} 1 & j<L-l, \\ 0, & otherwise, \end{cases}
\end{equation}
where $l$ is the predicted length. This mask offers flexibility across different stages of model development and application: during pre-training, a random $l$ exposes the model to various forecasting horizons, while in fine-tuning and inference, a fixed $l$ aligns with specific task requirements.
Moreover, stride mask $\mathbf{M}^{\text{S}}$, a variant of $\mathbf{M}^{\text{B}}$,  is designed for intermittent placement within time series during task-agnostic pretraining:
\begin{equation}
    \mathbf{M}^{\text{S}}(x, n_{\text{blocks}}) = 
\begin{cases} 
1 & \left\lfloor \frac{j}{b} \right\rfloor \bmod 2 = 0 \\ 
0 & \text{otherwise},
\end{cases}
\end{equation}
where $n_{block}$ is the number of blocks; \(b = \left\lceil \frac{L}{n_{\text{blocks}}} \right\rceil\) is the length of each block; \(j\) is the index of the sequence. It improves the modeling of temporal and inter-correlated dependencies by integrating information across non-contiguous parts of time series, leveraging neighboring values as additional context. 
In addition, reconstruction mask $\mathbf{M}^{\text{Rec}} = 0$ is employed for tasks such as synthetic data generation and anomaly detection. It allows direct generation of synthetic data or calculation of anomaly scores for each temporal position by comparing original and reconstructed series.

\subsection{Physics-Informed  Sampling}
\begin{algorithm}[t]
\caption{Physics-Informed TimeDiT through Energy-based Sampling}
\label{alg:energy_sampling}
\small
\begin{algorithmic}[1]
    \State $\bm{x}_T \sim \mathcal{N}(\bm{0}, \bm{I})$
    \For{$t=T, \dotsc, 1$}
        \State $\bm{z} \sim \mathcal{N}(\bm{0}, \bm{I})$ if $t > 1$, else $\bm{z} = \bm{0}$
        \State $\bm{x}_{t-1} = \frac{1}{\sqrt{\alpha_t}}\left( \bm{x}_t - \frac{1-\alpha_t}{\sqrt{1-\bar\alpha_t}} \bm{\epsilon}_\theta(\bm{x}_t, t) \right) + \sigma_t \bm{z}$
    \EndFor
    
    \For{$j=0,1,\dotsc,k-1$}
        \State $\bm{x}^{tar}_{j+1}=\bm{x}^{tar}_j+\epsilon\nabla K(\bm{x}_j^{tar};\bm{x}^{obs})+\alpha\epsilon\nabla\log p(\bm{x}_j^{tar}|\bm{x}^{obs}) + \sqrt{2\epsilon}\sigma, \sigma \sim \mathcal{N}(0,1)$
    \EndFor
    \State \textbf{return} $\bm{x}^{tar}_k$
\end{algorithmic}
\end{algorithm}

Physics principles are fundamental in shaping the evolution of temporal signals observed in real-world phenomena, such as climate patterns and oceanographic data. Therefore, it is essential to integrate physical knowledge into foundational time series models. In this section, we aim at developing a decoding method that can ensure the $\mathbf{x}^\text{{tar}}$ generated by \modelnamespace to satisfy our prior knowledge to the physical laws. To this end, we propose a strategy to incorporate physics knowledge as an energy-based prior for \modelnamespace during inference, which iteratively refines the reverse diffusion process. By guiding the denoising process during inference with gradients derived from physical laws represented by partial differential equations ({PDEs}), the integration of this knowledge can ensure $\mathbf{x}^\text{{tar}}$ to satisfy the PDEs and significantly enhance the quality of the generated samples.

We first start with a brief introduction to physical laws and PDE. A generic form of a physical law represented as a PDE that describes the evolution of a continuous temporal signal $\mathbf{x}(\mathbf{u},t)$ over a spatial coordinate $\mathbf{u}$ is given by:
\begin{equation}
    \label{eq:generic-pde}
     \frac{\partial \mathbf{x}}{\partial t} = F(t,\mathbf{x},\mathbf{u},\frac{\partial \mathbf{x}}{\partial \mathbf{u}_i}, \frac{\partial^2\mathbf{x}}{\partial \mathbf{u}_i \partial \mathbf{u}_j}, \dots)
\end{equation}

Based on this PDE representation of physical knowledge, the consistency between the predicted time series $\mathbf{x}^\text{{tar}}$ and the physics knowledge can be quantified using the following squared residual function:
\begin{equation}
    K(\mathbf{x}^\text{{tar}};F) = -||\frac{\partial \mathbf{x}^\text{{tar}}}{\partial t}-F(t,\mathbf{x}^\text{{tar}},\mathbf{u},\frac{\partial \mathbf{x}^\text{{tar}}}{\partial \mathbf{u}_i}, \frac{\partial^2\mathbf{x}^\text{{tar}}}{\partial \mathbf{u}_i \partial \mathbf{u}_j}, \dots)||^2_2
\end{equation}

This function reaches its maximum when the predicted time series is perfectly consistent with the physical model, resulting in a residual of 0. Using this metric $K$, physics knowledge can be integrated into a probabilistic time series foundation model $p(\mathbf{x}^\text{{tar}}|\mathbf{x}^\text{{con}})$ as an explicit regularization by solving the following optimization problem to obtain a refined model $q(\mathbf{x}^\text{{tar}}|\mathbf{x}^\text{{con}})$: 
   \small\begin{equation}
\label{eqx}
\begin{split}
q(\mathbf{x}^\text{tar}|\mathbf{x}^\text{con}) = \arg\max_{q} \Big[  \mathbb{E}_{\mathbf{x}^\text{tar}\sim q}K(\mathbf{x}^\text{tar};F) - \alpha D_{KL}(q(\mathbf{x}^\text{tar}|\mathbf{x}^\text{con})||p(\mathbf{x}^\text{tar}|\mathbf{x}^\text{con})) \Big]
\end{split}
\end{equation}
where the first term represents the aforementioned physics knowledge metric, and the second term controls the divergence between $q(\mathbf{x}^\text{{tar}}|\mathbf{x}^\text{{con}})$ and $p(\mathbf{x}^\text{{tar}}|\mathbf{x}^\text{{con}})$. However directly updating the model parameters to optimize the above function is resource-consuming. To solve this issue, we derived the closed-form solution, which does not need updating the model parameters. 
The above optimization problem has a closed-form solution as provided by the following theorem:

\begin{theorem}\label{th1}
    The optimal $q(\mathbf{x}^\text{tar}|\mathbf{x}^\text{{con}})$ in Eq.\ref{eqx} is the Boltzmann distribution defined on the following energy function:
    \begin{equation}
        \label{eq7}E(\mathbf{x}^{tar};\mathbf{x}^{con}) = K(\mathbf{x}^{tar};F)+\alpha\log p(\mathbf{x}^{tar}|\mathbf{x}^{con})
    \end{equation}
    in other words, the optimal $q(\mathbf{x}^{tar}|\mathbf{x}^{con})$ is:
    \begin{equation}
        q(\mathbf{x}^{tar}|\mathbf{x}^{con}) = \frac{1}{Z}\exp(K(\mathbf{x}^{tar};F)+\alpha\log p(\mathbf{x}^{tar}|\mathbf{x}^{con})),
    \end{equation}
where $Z = \int \exp(K(\mathbf{x}^{tar};F)+\alpha\log p(\mathbf{x}^{tar}|\mathbf{x}^{con}))d\mathbf{x}^{tar}$ is the partition function.

\end{theorem}
The theorem illustrates that sampling from the Boltzmann distribution defined in Eq. ~\ref{eq7}, is analogous to incorporating physics knowledge into model edition. In the context of diffusion models, this distribution can be effectively sampled using Langevin dynamics~\citep{stoltz2010free}:
\begin{equation}
    \begin{aligned}
        \mathbf{x}^\text{{tar}}_{j+1} &= \mathbf{x}^\text{{tar}}_j + \epsilon\nabla \log q(\mathbf{x}^\text{{tar}}|\mathbf{x}^\text{{con}}) + \sqrt{2\epsilon}\sigma, \sigma \sim \mathcal{N}(0,1)\\
        &=\mathbf{x}^\text{{tar}}_j+\epsilon\nabla K(\mathbf{x}^\text{{tar}}_j;\mathbf{x}^\text{{con}})+\alpha\epsilon\nabla\log p(\mathbf{x}^\text{{tar}}_j|\mathbf{x}^\text{{con}}) + \sqrt{2\epsilon}\sigma, \sigma \sim \mathcal{N}(0,1)
    \end{aligned}
    \end{equation}
where $\sigma \sim \mathcal{N}(0,1)$. In diffusion model, precisely calculate the likelihood $\log p(\mathbf{x}^\text{{tar}}|\mathbf{x}^\text{{con}})$ is intractable. To tackle this issue, following previous works \citep{kollovieh2023predict}, we approximate likelihood with the objective to edit the pre-trained diffusion model: $\log p(\mathbf{x}^\text{{tar}}|\mathbf{x}^\text{{con}}) = -\mathbb{E}_{\epsilon, t} [||\epsilon_\theta(\mathbf{x}^\text{tar},t;\mathbf{x}^\text{{con}})-\epsilon||^2]$.
The approximation presented above constitutes the optimizable component of the evidence lower bound(ELBO). Algorithm \ref{alg:energy_sampling} summarizes the comprehensive model editing process.

\begin{table}[!ht]
\centering
\caption{Full results of long sequence forecasting experiments evaluated on diverse datasets, where \textbf{bold} indicates the best performance and \underline{underlined} indicates the second best performance. Baseline results are sourced from \cite{woo2024unified}.}
\resizebox{\linewidth}{!}{
\begin{tabular}{crccccccccccccccccccc}
\toprule
&    & \multicolumn{4}{c}{\textbf{Zero-shot}}                 & \multicolumn{8}{c}{\textbf{Full-shot}} \\
\cmidrule(lr){3-6} \cmidrule(lr){7-14}
&  & \textbf{TimeDiT} & \textbf{\smallmodel} & \textbf{\basemodel} & \textbf{\largemodel} & \textbf{iTransformer} & \textbf{TimesNet} & \textbf{PatchTST} & \textbf{Crossformer} & \textbf{TiDE} & \textbf{DLinear} & \textbf{SCINet} & \textbf{FEDformer} \\
\midrule
\multirow{5}[1]{*}{\shortstack{\textbf{ETTh1}\\ \textbf{(MSE)}}}
& 96  & \textbf{0.325} & \underline{0.375} & 0.384 & 0.380 & 0.386 & 0.384 & 0.414 & 0.423 & 0.479 & 0.386 & 0.654 & \underline{0.376} \\
& 192 & \textbf{0.347} & \underline{0.399} & 0.425 & 0.440 & 0.441 & 0.436 & 0.460 & 0.471 & 0.525 & 0.437 & 0.719 & 0.420 \\
& 336 & \textbf{0.347} & \underline{0.412} & 0.456 & 0.514 & 0.487 & 0.491 & 0.501 & 0.570 & 0.565 & 0.481 & 0.778 & 0.459 \\
& 720 & \textbf{0.404} & \underline{0.413} & 0.470 & 0.705 & 0.503 & 0.521 & 0.500 & 0.653 & 0.594 & 0.519 & 0.836 & 0.506 \\\cline{2-14}
&Avg. & \textbf{0.356} & \underline{0.400} & {0.434} & 0.510 & 0.454 & 0.458 & 0.469 & 0.529 & 0.541 & 0.456 & 0.747 & 0.440 \\
\midrule
\multirow{5}[0]{*}{\shortstack{\textbf{ETTh2}\\ \textbf{(MSE)}} }
& 96  & \textbf{0.257} & {0.281} & \underline{0.277} & 0.287 & 0.297 & 0.340 & 0.302 & 0.745 & 0.400 & 0.333 & 0.707 & 0.358 \\
& 192 & \textbf{0.316} & \underline{0.340} & \underline{0.340} & 0.347 & 0.380 & 0.402 & 0.388 & 0.877 & 0.528 & 0.477 & 0.860 & 0.429 \\
& 336 & \textbf{0.341} & \underline{0.362} & {0.371} & 0.377 & 0.428 & 0.452 & 0.426 & 1.043 & 0.643 & 0.594 & 1.000 & 0.496 \\
& 720 & 0.447 & \textbf{0.380} & \underline{0.394} & 0.404 & 0.427 & 0.462 & 0.431 & 1.104 & 0.874 & 0.831 & 1.249 & 0.463 \\\cline{2-14}
&Avg. & \textbf{0.340}& \underline{0.341} & {0.345} & 0.354 & 0.383 & 0.414 & 0.387 & 0.942 & 0.611 & 0.559 & 0.954 & 0.437 \\
\midrule
\multirow{5}[1]{*}{\shortstack{\textbf{ETTh1}\\ \textbf{(MAE)}}} 
& 96  & \textbf{0.386} & 0.402 & 0.402 & \underline{0.398} & 0.405 & 0.402 & 0.419 & 0.448 & 0.464 & 0.400 & 0.599 & \underline{0.419} \\
& 192 & \textbf{0.390} & \underline{0.419} & 0.429 & 0.434 & 0.436 & 0.429 & 0.445 & 0.474 & 0.492 & 0.432 & 0.631 & 0.448 \\
& 336 & \textbf{0.396} & \underline{0.429} & 0.450 & 0.474 & 0.458 & 0.469 & 0.466 & 0.546 & 0.515 & 0.459 & 0.659 & 0.465 \\
& 720 & \textbf{0.440} & \underline{0.444} & 0.473 & 0.568 & 0.491 & 0.500 & 0.488 & 0.621 & 0.558 & 0.516 & 0.699 & 0.507 \\\cline{2-14}
&Avg. & \textbf{0.403} & \underline{0.424} & {0.438} & 0.469 & 0.448 & 0.450 & 0.455 & 0.522 & 0.507 & 0.452 & 0.647 & 0.46 \\
\midrule
\multirow{5}[0]{*}{\shortstack{\textbf{ETTh2}\\ \textbf{(MAE)}}} 
& 96  & 0.331 & 0.334 & \underline{0.327} & \textbf{0.325} & 0.349 & 0.374 & 0.348 & 0.584 & 0.440 & 0.387 & 0.621 & 0.397 \\
& 192 & 0.380 & \underline{0.373} & 0.374 & \textbf{0.367} & 0.400 & 0.414 & 0.400 & 0.656 & 0.509 & 0.476 & 0.689 & 0.439 \\
& 336 & 0.418 & \underline{0.393} & 0.401 & \textbf{0.393} & 0.432 & 0.541 & 0.433 & 0.731 & 0.571 & 0.541 & 0.744 & 0.487 \\
& 720 & 0.441 & \textbf{0.416} & 0.426 & \underline{0.421} & 0.445 & 0.657 & 0.446 & 0.763 & 0.679 & 0.657 & 0.838 & 0.474 \\\cline{2-14}
 &Avg.& 0.393 & \underline{0.379} & 0.382 & \textbf{0.376} & 0.407 & 0.497 & 0.407 & 0.684 & 0.550 & 0.515 & 0.723 & 0.449 \\
\bottomrule
\end{tabular}%
}
\label{tab:lsf_full}
\end{table}

\begin{table}[!ht]
\centering
\caption{Forecasting results on CRPS\_{sum} for both zero-shot and full-shot settings.}
\resizebox{0.9\textwidth}{!}{
\begin{tabular}{ccccccccccc}\toprule
\textbf{Setting} &\textbf{Dataset}  & \textbf{TimeDiT (ZS)} & \textbf{TEMPO} & \textbf{\smallmodel} & \textbf{\basemodel} & \textbf{\largemodel} & \textbf{LagLLaMA} & \textbf{TimeMixer} & \textbf{TimeLLM} & \textbf{Timer} \\ \toprule
\multirow{5}[1]{*}{\textbf{Zero Shot}} &\textbf{Solar} &  \textbf{0.424} & \underline{0.581} & 0.884 & 0.948 & 1.042 & 0.690 & 0.999&0.997&0.101 \\
&\textbf{Electricity} &  \textbf{0.030} & 0.081 & 0.079 & 0.072 & \underline{0.039} & 0.065 &0.302&0.303&0.301\\
&\textbf{Traffic} &  0.351 & \underline{0.147} & 0.215 & 0.191 & \textbf{0.111} & 0.275 &0.403&0.368&0.384 \\
&\textbf{Taxi} &  \textbf{0.392} & \underline{0.400} & 0.463 & 0.428 & 0.597 & 0.620&0.785 &0.782 &0.788 \\
&\textbf{Exchange} &  0.019 & 0.030 & \textbf{0.007} & 0.012 & \underline{0.011} & 0.024 &0.079 &0.076 &0.072 \\\toprule
\textbf{Setting} &\textbf{Dataset} & \textbf{TimeDiT (FS)} & \textbf{DLinear} & \textbf{PatchTST} & \textbf{Latent ODE} & \textbf{GPT4TS} &  \textbf{TransMAF} & \textbf{TimeGrad} & \textbf{CSDI} & \textbf{Diffusion-TS} \\ \midrule
\multirow{5}[1]{*}{\textbf{Full Shot}} &\textbf{Solar}          & \textbf{0.278} & 0.432 & 0.457 & 0.445 & 0.467 & 0.301 & 0.287 & 0.298 & \underline{0.286} \\
&\textbf{Electricity}    & \textbf{0.005} & 0.033 & 0.037 & 0.140 & 0.033 &  0.021 & 0.021 & \underline{0.017} & 0.019 \\
&\textbf{Traffic}        & \textbf{0.019} & 0.070 & 0.405 & 0.095 & 0.069 & 0.056 & 0.044 & \underline{0.020} & 0.097 \\
&\textbf{Taxi}           & \underline{0.123} & 0.177 & 0.190 & 0.181 & 0.187 &  0.179 & \textbf{0.114} & 0.123 & 0.303 \\
&\textbf{Exchange}       & \textbf{0.005} & 0.011 & 0.026 & 0.013 & 0.013 &  \underline{0.005} & 0.006 & 0.007 & 0.009 \\ 

\bottomrule
\end{tabular}}
\label{combined_forecasting}
\end{table}
\section{Experiments}

The development of foundation models for time series analysis presents unique challenges at the intersection of universal representation learning and task-specific adaptation. While traditional foundation models in domains like natural language processing demonstrate strong zero-shot capabilities, time series tasks often require nuanced understanding of domain-specific characteristics despite sharing fundamental temporal dependencies. This observation motivated our comprehensive three-perspective experimental framework: First, we assess the model's foundational capabilities through both zero-shot transfer and fine-tuning approaches on canonical tasks, including forecasting and imputation on ETTh, Solar, Electricity, Traffic, Taxi, Exchange, ETTm, Weather. Second, we investigate domain-adaptive tasks that require prior knowledge of data characteristics—specifically miss-value and multi-resolution forecasting including Air Quality from climate, MIMIC-III and PhysioNet from healthcare, and NASDAQ from finance, anomaly detection on MSL, SMAP, SWaT, SMD, and PSM datasets\citep{xu2021anomaly, zhao2020multivariate}, and targeted data generation on Stock, Air Quality, and Energy datasets\citep{yoon2019time, desai2021timevae}. These experiments evaluate the model's capability to bridge the gap between universal temporal patterns and domain-specific requirements. Finally, we examine the model's capacity for knowledge integration through physics-informed constraints by accurately processing complex partial differential equations (PDEs) \citep{yuan2024diffusion} and multimodal fusion, testing its ability to incorporate external domain knowledge to enhance its foundational capabilities further.

\subsection{Zero-Shot and Finetuning Setting}
\label{zeroshot}
\paragraph{Forecasting on Zero-shot Setting:}

We comprehensively evaluate TimeDiT in zero-shot forecasting environments against leading architectures, encompassing both zero-shot and full-shot models. Using popular benchmark datasets for long-term deterministic forecasting that exclude source overlap with pre-training data, TimeDiT demonstrates exceptional performance across varying prediction horizons (96-720 time steps) as shown in  Table~\ref{tab:lsf_full}. The model achieves superior Mean Squared Error (MSE) and Mean Absolute Error (MAE) metrics compared to all baselines, particularly excelling on the ETTh1 dataset where it consistently outperforms all comparative models.

Building upon this success, we extended our investigation to probabilistic forecasting, where TimeDiT demonstrates remarkable capabilities. As evidenced in Table~\ref{combined_forecasting}, zero-shot TimeDiT achieves superior performance compared to models with fewer parameters in equivalent zero-shot settings. We further fine-tuned TimeDiT, motivated by both its strong zero-shot performance and the inherent challenges in probabilistic modeling, where models typically struggle with calibration issues—either producing overconfident predictions with overly narrow distributions or underconfident predictions with excessively wide intervals due to insufficient domain-specific knowledge. The CRPS\_sum evaluation reveals that fine-tuned TimeDiT surpasses specialized probabilistic forecasting models across multiple datasets, demonstrating exceptional performance in both zero-shot generalization and fine-tuned specialization. These results establish TimeDiT as a significant advancement in time series modeling, particularly for applications requiring robust uncertainty quantification.

\begin{table}[!t]
\centering
\caption{Imputation results on multivariate time series averaged over the four mask ratios.}
\resizebox{0.7\textwidth}{!}{
\begin{tabular}{clcccccc}
\toprule
Metrics&\textbf{Datasets} & \textbf{ETTh1} & \textbf{ETTh2} & \textbf{ETTm1} & \textbf{ETTm2} & \textbf{Weather} & \textbf{Electricity} \\
\midrule
\multirow{7}[1]{*}{\textbf{MSE}} &\textbf{PatchTST}        & 0.115 & 0.065 & 0.047 & 0.029 & 0.034 & \underline{0.072} \\
&\textbf{TimesNet}        & 0.078 & 0.049 & \underline{0.027} & \underline{0.022} & \textbf{0.030} & 0.092 \\
&\textbf{GPT4TS}          & \underline{0.069} & \underline{0.048} & 0.028 & \textbf{0.021} & \underline{0.031} & 0.090 \\
&\textbf{Timer}           & 0.145 & 0.077 & 0.051 & 0.035 & 0.108 & 0.097 \\
&\textbf{TimeMixer}       & 0.119 & 0.064 & 0.051 & 0.028 & 0.031 & 0.061 \\
&\textbf{iTransformer}    & 0.149 & 0.150 & 0.071 & 0.083 & 0.053 & 0.099 \\
\cline{2-8}
& \textbf{\modelname}      & \textbf{0.036} & \textbf{0.031} & \textbf{0.022} & 0.034 & \underline{0.031} & \textbf{0.068} \\
\midrule
\multirow{7}[1]{*}{\textbf{MAE}}&\textbf{PatchTST}        & 0.224 & 0.163 & 0.140 & 0.102 & 0.055 & \underline{0.183} \\
&\textbf{TimesNet}        & 0.187 & 0.146 & 0.107 & \underline{0.088} & \underline{0.054} & 0.210 \\
&\textbf{GPT4TS}          & \underline{0.173} & \underline{0.141} & \underline{0.105} & \textbf{0.084} & 0.056 & 0.207 \\
&\textbf{Timer}           & 0.243 & 0.172 & 0.141 & 0.105 & 0.168 & 0.194 \\
&\textbf{TimeMixer}       & 0.226 & 0.157 & 0.143 & 0.093 & 0.049 & 0.164 \\
&\textbf{iTransformer}    & 0.270 & 0.271 & 0.185 & 0.192 & 0.116 & 0.224 \\
\cline{2-8}
&\textbf{\modelname}      & \textbf{0.122} & \textbf{0.111} & \textbf{0.093} & 0.104 & \textbf{0.036} & \textbf{0.172} \\
\bottomrule
\end{tabular}}
\label{tab:imputation}
\end{table}

\paragraph{Imputation Task:}
Leveraging our model's strong temporal understanding from forecasting tasks, rather than training a separate model or modifying the architecture, we strategically opt to fine-tune the same pre-trained checkpoint used in our forecasting experiments, maintaining architectural consistency while adapting the model's capabilities to the imputation task.
We conduct experiments on six benchmark time-series datasets: ETTh1, ETTh2, ETTm1, ETTm2, Electricity, and Weather. We use random mask ratios $\{12.5\%, 25\%, 37.5\%, 50\%\}$ following previous studies' settings with sequence length set to 96.
Table \ref{tab:imputation} shows the imputation result averaged over the four mask ratios. 
\modelnamespace demonstrates superior performance, achieving the best results in 10 out of 12 evaluations, while all other baselines combined secured only 2 top positions. Notably,  \modelnamespace achieved a 39\% reduction in MSE and 22\% reduction in MAE compared to the strongest baseline on the ETTh1 dataset. For full result on each mask ratio, please refer to section \ref{impute_details}.

\begin{wrapfigure}{l}{0.45\textwidth}
    \centering
    \vspace{-.1in}
    \includegraphics[height= 3.5cm]{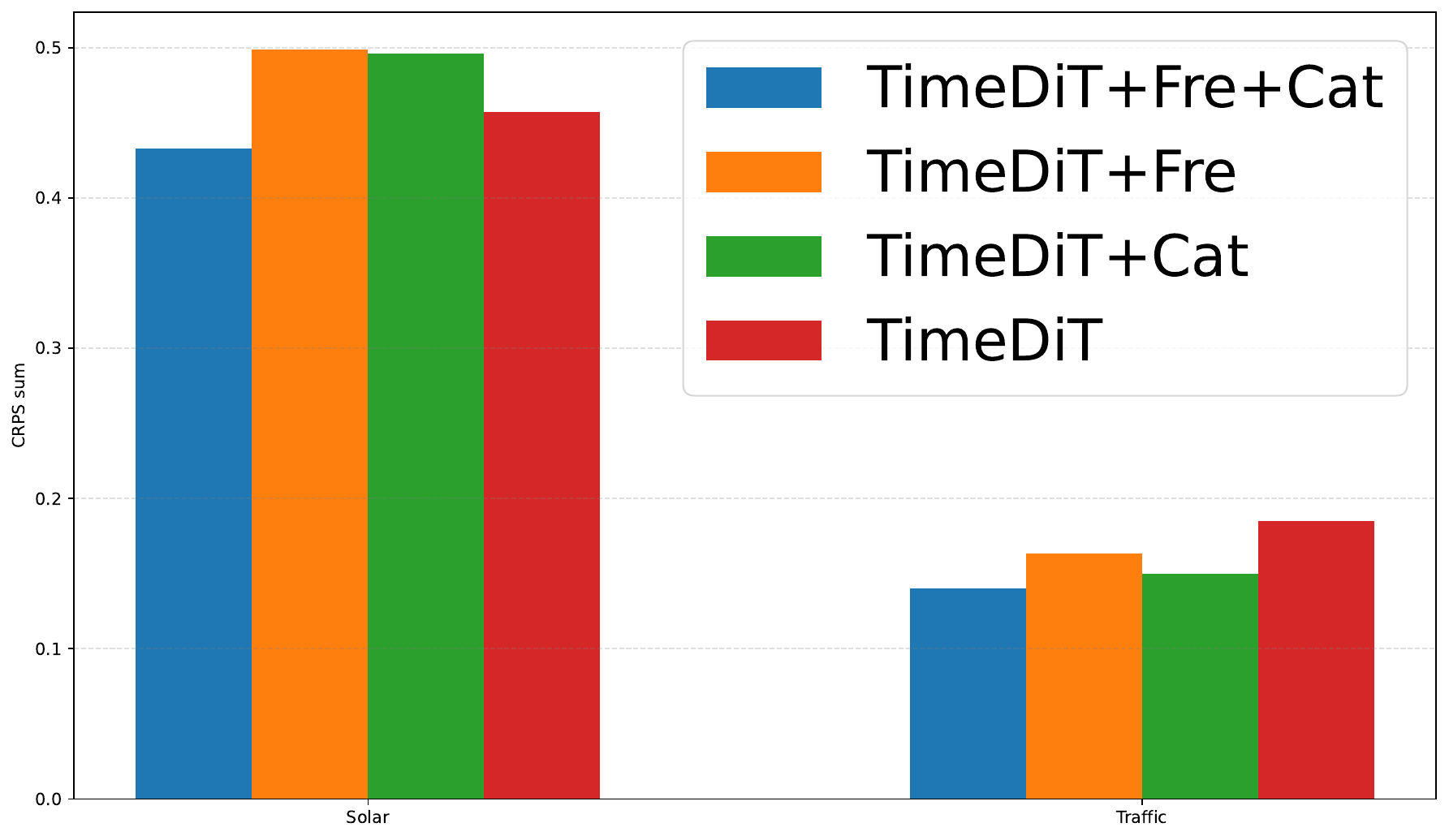}
    \caption{TimeDiT with textual information.}
    \label{multimodal_pic}
    \vspace{-.1in}
\end{wrapfigure}
\paragraph{Multimodal TimeDiT:}
While textual information is intuitively crucial for precise time series analysis, effectively aligning textual and numerical data has remained challenging. To address this, we explore the integration of textual information as classifiers in TimeDiT, incorporating two key elements as guidance ($c$ in Figure~\ref{fig:model_overview}): TSD's frequency (Fre) for capturing temporal periodicity, and TSD's categories (Cat) for representing domain-specific features. We pre-train three variants of TimeDiT and apply them in a zero-shot setting on Solar and Traffic datasets. The results in Figure~\ref{multimodal_pic} demonstrate that utilizing both types of information significantly boosts zero-shot performance, indicating TimeDiT's capacity to leverage external information for rapid adaptation to both learned and specific representations. Comparing single-term guidance with the combined TimeDiT+Fre+Cat model reveals that precise, multi-faceted information is necessary to achieve optimal results. These experiments highlight that TimeDiT's integration of textual context improves forecasting accuracy, enabling more informed decision-making in real-world time series applications.

\begin{table}[t]
\label{tab.physics}
\renewcommand{\arraystretch}{1.2}
\centering
{

\caption{Physics-informed TimeDiT results for PDE forecasting, including both mean error and error bars. Lower values indicate better performance and closer adherence to physical laws.}
\resizebox{0.9\textwidth}{!}{
\begin{tabular}{lcccc|cccc}
\toprule

       ~& \textbf{MSE} & \textbf{RMSE} & \textbf{MAE} & \textbf{CRPS} & \textbf{MSE} & \textbf{RMSE} & \textbf{MAE} & \textbf{CRPS}\\ \hline
        ~ & \multicolumn{4}{c}{Advection} & \multicolumn{4}{c}{Navier-Stokes} \\ \hline
        DDPM & \underline{0.011(0.000)} & \underline{0.106(0.001)} & \underline{0.084(0.001)} & \underline{0.472(0.007)} & \underline{0.309(0.004)} & \underline{0.556(0.004)} & 0.332(0.005) & 0.415(0.006) \\ 
        DDIM & 0.015(0.000) & 0.122(0.002) & 0.096(0.001) & 0.559(0.009) & 0.350(0.014) & 0.591(0.011) & 0.377(0.009) & 0.470(0.013) \\ 
        TSDiff & \underline{0.011(0.000)} & 0.106(0.022) & 0.085(0.001) & 0.472(0.011) & 0.399(0.008) & 0.556(0.007) & \underline{0.331(0.006)} & \underline{0.414(0.007)} \\
        \modelname & \textbf{0.010(0.000)} & \textbf{0.103(0.002)} & \textbf{0.082(0.001)} & \textbf{0.464(0.008)} & \textbf{0.299(0.006)} & \textbf{0.546(0.006)} & \textbf{0.322(0.06)} & \textbf{0.403(0.007)} \\ \hline
        ~ & \multicolumn{4}{c}{Burgers} & \multicolumn{4}{c}{Vorticity} \\ \hline
        DDPM & \underline{0.016(0.001)} & \underline{0.128(0.004)} & \underline{0.101(0.003)} & \underline{1.787(0.040)} & 1.917(0.020) & 1.385(0.007) & 0.851(0.009) & 0.476(0.005) \\ 
        DDIM & 0.018(0.000) & 0.136(0.001) & 0.116(0.001) & 1.858(0.015) & \underline{1.567(0.031)} & \underline{1.252(0.012)} & \textbf{0.754(0.012)} & \textbf{0.401(0.006)} \\ 
        TSDiff & 0.017(0.001) & 0.129(0.005) & 0.102(0.004) & 1.800(0.055) & 1.966(0.073) & 1.402(0.026) & 0.866(0.010) & 0.485(0.005) \\ 
        \modelname & \textbf{0.011(0.001)} & \textbf{0.104(0.005)} & \textbf{0.083(0.003)} & \textbf{1.395(0.053)} & \textbf{1.524(0.523)} & \textbf{1.234(0.021)} & \underline{0.772(0.009)} & \underline{0.445(0.006)} \\ \hline
         ~ & \multicolumn{4}{c}{Diffusion Sorption} & \multicolumn{4}{c}{CFD} \\ \hline
        DDPM & \underline{0.309(0.004)} & \underline{0.556(0.004)} & {0.332(0.005)} & \underline{0.415(0.006)} & \underline{0.004(0.000)} & \underline{0.065(0.001)} & \underline{0.054(0.000)} & \underline{0.082(0.000)} \\ 
        DDIM & 0.349(0.013) & 0.591(0.011) & 0.377(0.009) & 0.470(0.013) & {0.039(0.002)} & {0.194(0.006)} & {0.188(0.006)} & {0.313(0.012)} \\ 
        TSDiff & 0.309(0.008) & 0.556(0.007) & \underline{0.331(0.006)} & \textbf{0.414(0.007)} & N/A & N/A & N/A & N/A \\ 
        \modelname & \textbf{0.284(0.005)} & \textbf{0.533(0.005)} & \textbf{0.327(0.005)} & {0.423(0.007)} & \textbf{0.004(0.000)} & \textbf{0.062(0.001)} & \textbf{0.051(0.001)} & \textbf{0.080(0.001)} \\
        \bottomrule
\end{tabular}}
\label{tab:physics_mse_rmse_mae_crps}
}
\end{table}

\subsection{Physics-Informed TimeDiT} 

Our approach enables the direct incorporation of physics knowledge into the pre-trained foundation model without fine-tuning.
In this section, we evaluate how effectively our pre-trained foundation model can integrate physics-informed knowledge into time series forecasting.
We study six 1D partial differential equations (PDEs) forecasting from ~\citep{takamoto2022pdebench}: general Navier-Stokes Equations, Kolmogorov Flow (a specific case of Navier-Stokes Equations), Advection Equations, Burgers Equations, Diffusion Soeption and Computational Fluid Dynamics (CFD). 
Table~\ref{tab:physics_mse_rmse_mae_crps}  clearly demonstrates that our proposed model editing solution, which incorporates physics knowledge, significantly outperforms previous sampling strategies {introduced in }
DDPM~\citep{ho2020denoising}, DDIM~\citep{songdenoising}, and TS Diffusion, which proposes the Self-Guidance~\citep{kollovieh2023predict} to improve sampling quality. By leveraging domain-specific physical information, our approach achieves substantial performance improvements over these baselines, highlighting the effectiveness of integrating physics-informed priors into the diffusion model sampling process. This represents a novel advance in scientific machine learning, enabling rapid adaptation to specific physical systems.

\subsection{In Domain Setting}

\begin{table}[H]
\centering
\caption{Anomaly Detection result on 100-length multivariate time series. We calculate F1 score as \% for each dataset. '$.$' notation in model name stands for transformer.}
\resizebox{0.65\textwidth}{!}{
\begin{tabular}{lc|ccccc}
\toprule
\textbf{Methods} & \textbf{\modelname} &{\textbf{TimeMixer}}&{\textbf{iTrans.}}&\textbf{GPT4TS} & \textbf{TimesNet} & \textbf{PatchTS.} \\
\midrule
MSL & \textbf{89.33} & 81.95 & 72.54 & 82.45 & 81.84 & 78.70 \\
SMAP & \textbf{95.91} & 67.63 & 66.76 & \underline{72.88} & 69.39 & 68.82 \\
SWaT & \textbf{96.46} & 88.84 & 92.63 & \underline{94.23} & 93.02 & 85.72 \\
SMD & 83.28 & 78.33 & 82.08 & \textbf{86.89} & 84.61 & 84.62 \\
PSM & \textbf{97.57} & 93.11 & 95.32 & 97.13 & \underline{97.34} & 96.08 \\
\midrule
1st Pl Count & 4 & 0 & 0 & 1 & 0 & 0 \\
\bottomrule
\end{tabular}}
\label{tab:ad_main}
\end{table}

\paragraph{Anomaly Detection Task:}

We conduct experiments on five real-world datasets from industrial applications: MSL, SMAP, SWaT, SMD, and PSM. Given the distinct characteristics and patterns of anomalies in different industrial systems, in-domain evaluation is crucial for assessing model robustness and reliability in practical settings.
As shown in Table \ref{tab:ad_main}, \modelnamespace outperforms baseline models on four of the five datasets. Notably, on the SMAP dataset, \modelnamespace achieves a remarkable 23.03-point improvement in F1 score compared to the previous best baseline. These results demonstrate the effectiveness of our approach in handling real-world anomaly detection scenarios across various industrial applications.

\begin{table}[t]
\centering
\caption{Synthetic Generation results on 24-length multivariate time series. We calculate discriminative and predictive scores according to \citep{yoon2019time}.} 
\scalebox{0.8}{
\begin{tabular}{cccccc}
\toprule
\textbf{Metric} & \textbf{Methods} & \textbf{Sine} & \textbf{Stocks} & \textbf{Air Quality}  & \textbf{Energy} \\
\midrule
\multirow{4}{*}{\shortstack{Discriminative \\ Score}} 
& TimeGAN & 0.1217(0.039) & 0.2038(0.057) & 0.3913(0.039) & 0.4969 (0.000)  \\
& TimeVAE & 0.0489(0.0562) & 0.1987(0.037) & 0.2869(0.053) & 0.4993(0.001) \\
&Diffusion-TS & \underline{0.0099(0.003)}& \underline{0.1869(0.0159)}& \textbf{0.1227(0.006)} & \underline{0.2301(0.006)} \\
& \modelname & \textbf{0.0086(0.004)} & \textbf{0.0087(0.006)} & \underline{0.1923(0.003)}& \textbf{0.0053(0.002)} \\
\midrule
\multirow{4}{*}{\shortstack{Predictive \\ Score}} 
& TimeGAN & 0.2797(0.015) & 0.0481(0.002) & 0.035(0.002) & 0.3305(0.003)\\
& TimeVAE & 0.2285(0.000) & 0.0485(0.000) & 0.0269(0.001) & 0.2878(0.001)  \\
& Diffusion-TS & \underline{0.2262(0.000)} & \textbf{0.042(0.000)} & \underline{0.022(0.002)} & \underline{0.2506(0.000)} \\
& \modelname & \textbf{0.1915(0.000)} & \underline{0.0445(0.000)} & \textbf{0.0217(0.000)} & \textbf{0.2489(0.000)} \\
\bottomrule
\end{tabular}}
\label{tab:syn}

\end{table}

\begin{table}[ht]
\centering
\caption{Forecasting results on practical scenarios with both deterministic metric (MAE/MSE) for \textit{accuracy} evaluation and probabilistic metric (CRPS/CRPS\_sum) for \textit{uncertainty quantification}. \textbf{Bold} indicates best result, \underline{Underline} indicates the second best result.}
\resizebox{0.7\textwidth}{!}{
\begin{tabular}{lcccccc}
\toprule
 & \textbf{Air Quality} & \textbf{MIMIC-III} & \textbf{PhysioNet(a)} & \textbf{PhysioNet(b)} & \textbf{PhysioNet(c)} & \textbf{NASDAQ} \\ \cline{2-7}
 & MAE/MSE & MAE/MSE & MAE/MSE & MAE/MSE & MAE/MSE & MAE/MSE \\
\hline
DLinear & 0.683/0.685 & 0.786/1.000 & 0.686/0.758 & 0.733/0.922 & 0.715/0.813 & 2.715/8.137 \\
Neural ODE & 0.678/0.679 & 0.784/0.999 & 0.685/0.756 & 0.732/0.918 & 0.713/0.811 & 3.227/11.155 \\
Neural CDE & 0.683/0.685 & 0.787/1.002 & 0.688/0.754 & 0.733/0.921 & 0.713/0.814 & 3.319/11.816 \\
PatchTST & 0.685/0.683 & 0.778/0.987 & 0.699/0.780 & 0.733/0.932 & 0.714/0.802 & 3.182/10.635 \\
GPT4TS & 0.696/0.701 & 0.750/0.921 & 0.697/0.772 & 0.734/0.921 & 0.713/0.817 & 3.176/10.873 \\
CSDI & 0.539/0.554 & \underline{0.551}/\underline{0.681} & \textbf{0.548}/\textbf{0.548} & \underline{0.665}/\underline{0.792} & \underline{0.665}/\underline{0.695} & \underline{0.524}/\textbf{0.388} \\
DiffTS & \underline{0.521}/\underline{0.538} & 0.677/0.908 & 0.610/0.742 & 0.701/0.880 & 0.678/0.872 & 1.951/9.515 \\
{TimeMixer} & 0.691/0.697 & 0.769/0.981 & 0.692/0.775 & 0.734/0.920 & 0.707/0.805 & 3.267/11.511 \\
{TimeLLM} & 0.701/0.705 & 0.787/1.020 & 0.687/0.761 & 0.731/0.931 & 0.713/0.800 & 3.125/10.276 \\
{MG-TSD} & 0.471/0.364 & - & - & - & - & 0.522/3.324 \\
TimeDiT & \textbf{0.457}/\textbf{0.354} & \textbf{0.517}/\textbf{0.534} & \underline{0.577}/\underline{0.620} & \textbf{0.659}/\textbf{0.766} & \textbf{0.543}/\textbf{0.561} & \textbf{0.516}/\underline{0.418} \\
\hline
\hline
 & CRPS/\_sum & CRPS/\_sum & CRPS/\_sum & CRPS/\_sum & CRPS/\_sum & CRPS \\
\hline
DLinear & 0.662/0.544 & 0.770/0.748 & 0.764/0.812 & 0.794/0.793 & 0.767/0.797 & 0.342 \\
Neural ODE & 0.657/0.529 & 0.769/0.733 & 0.763/0.806 & 0.792/0.789 & 0.765/0.793 & 0.426 \\
Neural CDE & 0.659/0.551 & 0.771/0.754 & 0.763/0.799 & 0.792/0.786 & 0.765/0.791 & 0.439 \\
PatchTST & 0.664/0.564 & 0.771/0.721 & 0.769/0.812 & 0.791/0.775 & 0.766/0.777 & 0.410 \\
GPT4TS & 0.666/0.584 & 0.751/0.690 & 0.767/0.809 & 0.795/0.798 & 0.770/0.768 & 0.419 \\
CSDI & \underline{0.598}/\underline{0.620}& \textbf{0.504}/0.798 & \underline{0.620}/\underline{0.641} & 0.725/0.787 & \underline{0.669}/0.748 & \underline{0.096} \\
DiffTS & 0.649/0.719 & 0.633/\underline{0.676} & 0.628/0.668 & \underline{0.720}/\underline{0.724} & 0.679/\underline{0.719} & 0.283 \\
{TimeMixer} & 0.667/0.576 & 0.776/0.724 & 0.763/0.805 & 0.794/0.798 & 0.757/0.784 & 0.432 \\
{TimeLLM} & 0.664/0.571 & 0.785/0.700 & 0.752/0.797 & 0.795/0.795 & 0.757/0.754 & 0.405 \\
{MG-TSD} & 0.579/0.564 & - & - & - & - & 0.275 \\
TimeDiT & \textbf{0.554}/\textbf{0.522} & \underline{0.599}/\textbf{0.649} & \textbf{0.616}/\textbf{0.640} & \textbf{0.708}/\textbf{0.710} & \textbf{0.668}/\textbf{0.708} & \textbf{0.091} \\ 
\bottomrule
\label{tab:1_MS_RES}
\end{tabular}}
\end{table}

\paragraph{Synthetic Generation Task:}

We conduct experiments to synthesize multivariate time series and evaluate performance using the discriminative score and predictive score metrics under a "train on synthetic test on real" experimental setup with sequence length set to 24 ~\citep{yuan2024diffusion}. 
Table \ref{tab:syn} shows the result on synthetic generation where TimeDiT, in general, consistently generates more realistic synthetic samples compared to baselines, even on challenging energy datasets. This demonstrates TimeDiT's strength in complex time series synthesis. 
PCA visualization of synthesis performance in Appendix~\ref{synthetic_visual} shows that TimeDiT's samples markedly overlap the original data distribution better than other methods. Qualitative and quantitative results confirm TimeDiT's superior ability to model intricate characteristics for realistic time series synthesis, even on multidimensional, complex datasets.

\vspace{-0.1in}
\paragraph{Practical Scenarios: Missing Data and Multi-Resolution Forecasting.}
To evaluate TimeDiT's performance in realistic scenarios, {we conducted experiments incorporating three real-world challenges: missing values {(validated on Air Quality, MIMIC)},  irregularly sampled time series~\citep{jeon2022gtgan, naiman2024generative} with varying time intervals between observations (evaluated on PhysioNet), multi-resolution data {(tested on NASDAQ)}.} This domain-specific testing ensures TimeDiT can handle both technical challenges (missing values, irregular sampling, multi-resolution) and domain-specific nuances, validating its practical utility across diverse applications. We evaluated forecasting accuracy using Mean Absolute Error (MAE) and Mean Squared Error (MSE), while uncertainty quantification (UQ) was assessed using Continuous Ranked Probability Score (CRPS) and CRPS\_sum. Results in Table~\ref{tab:1_MS_RES} demonstrate that TimeDiT not only achieves high accuracy in point forecasts but also provides well-calibrated probabilistic forecasts, effectively capturing the inherent uncertainties in complex time series data. The model's strong performance in probabilistic metrics indicates its ability to generate reliable prediction intervals and accurately represent the full predictive distribution. {This robust UQ capability, coupled with TimeDiT's ability to handle missing values and irregular samples without additional designs for interpolation, positions it as a powerful tool for decision-making in uncertain environments.}

\subsection{Ablation Study}
\vspace{-0.1in}
\begin{table}[ht]
\centering
\caption{Ablation study on zero-shot forecasting}
\label{main_ablation}
\resizebox{0.65\textwidth}{!}{\begin{tabular}{lccccc}
\toprule
Dataset & TimeDiT & w/o Random M. & w/o Stride M. & w/o Block M. & w/o Phys \\
\hline
Solar & \textbf{0.424} & {0.465} & 0.469 & 0.862& 0.445 \\
Electricity & \textbf{0.030} & 0.035 & 0.037 & 0.101 &0.033  \\
\hline\hline
Dataset &   Dual-attention & Channel-wise & Patch Token&Additive & Cross-attention \\
\hline
Solar &  0.467 & 0.461 & 0.874 &0.677&0.711\\
Electricity & 0.037& 0.039 & 0.145&0.079&0.077 \\\bottomrule
\end{tabular}}

\end{table}

As shown in Table~\ref{main_ablation}, we conduct comprehensive ablation studies to validate the effectiveness of TimeDiT's key components. The full TimeDiT model achieves superior performance (MSE of 0.424 for Solar and 0.030 for Electricity datasets) compared to its variants. Among the masking strategies, stride masking proves to be the most crucial component, as its removal leads to substantial performance degradation (MSE increasing to 0.862 for Solar and 0.101 for Electricity). In the attention mechanism analysis, our proposed temporal-attention architecture demonstrates consistent advantages over alternative designs, with patch token attention yielding the least favorable results (MSE of 0.874 and 0.145 for Solar and Electricity, respectively). Furthermore, the conditional AdaLN design emerges as the optimal choice among various conditioning approaches, reinforcing our architectural decisions.

\section{Conclusion}

In this paper, we introduce \modelname, a pioneering approach to creating a versatile and robust foundation model for various time series tasks under practical scenarios. By integrating transformer inductive bias with diffusion model, \modelnamespace effectively captures temporal dependencies and addresses real world challenges unique to time series regarding multi-resolution and missing values as well as incorporating external knowledge. Our innovative masking strategies allow for a consistent training framework adaptable to diverse tasks such as forecasting, imputation, and anomaly detection and synthetic data generation. 
We recognize some limitations of current work: first, we primarily explored common sequence lengths and did not assess \modelname's performance on very long sequences. While we have introduced randomness in prediction length and feature numbers up to a maximum, we aim to develop more scalable solutions for highly variable multivariate time series. Furthermore, our understanding of how different types of domain information contribute to performance improvement is still under investigation. In addition, we acknowledge the importance of sequence-level classification and are actively collecting datasets to extend TimeDiT's capabilities to classification tasks in future work. Lastly, there is a high demand for deeply developing foundation models for multi-modal time series, allowing \modelnamespace to utilize diverse data sources for enhanced performance.

\bibliography{ref}
\newpage
\appendix

\renewcommand{\appendixname}{Appendix}
\renewcommand{\appendixtocname}{Appendix}
\renewcommand{\appendixpagename}{Appendix}
\appendixpage

\startcontents[appendix]
\etocsetnexttocdepth{subsection}
\printcontents[appendix]{l}{1}{\setcounter{tocdepth}{3}}

\newpage

\section{TimeDiT Paradigm on Training and Inference}

{\paragraph{Position of TimeDiT} TimeDiT bridges the gap between ideal foundation models and current practical limitations in time series analysis. While existing approaches often struggle with task-specific architectures or limited domain adaptability, TimeDiT embodies key foundation model characteristics through a balanced design that integrates domain knowledge while maintaining versatile computational capabilities. Its foundation model status is established through three key aspects: (1) a unified mask mechanism that naturally accommodates varying channel sizes and sequence lengths, enabling processing of diverse time series data without task-specific architectures; (2) a versatile framework supporting multiple downstream tasks including forecasting, imputation, anomaly detection, and synthetic data generation; and (3) physics-informed sampling through an energy-based approach that incorporates domain knowledge during inference without model retraining. Although current technical constraints like sequence length limitations and multi-modal integration remain to be fully addressed, this combination of architectural flexibility, task-agnostic design, and domain knowledge integration positions TimeDiT as a practical proto-foundation model that advances the field toward universal time series modeling.
 }

\paragraph{Training details} Similar to the previous DiT work~\citep{Peebles2022DiT}, TimeDiT is available in four sizes: small (S, 33M parameters), big (B, 130M parameters), large (L, 460M parameters), and extra large (XL, 680M parameters). {
A comprehensive comparison in Table ~\ref{table:7_com} shows TimeDiT's expanded task coverage relative to existing general-purpose time series models, including anomaly detection, imputation and data generation.}
In our training process, we utilized the Adam optimizer with a learning rate of 0.0001 and the loss function is from Equantion~\ref{loss}. Batch sizes of 256 or 512 were employed, depending on model size. The ideal epoch to convergence is over 100 as the complexity of training data, but we choose to use the earlier checkpoint for the case of downstream purpose of anomaly detection and synthetic generation because the two tasks are very dataset-specific and do not necessarily benefit from learning distributions beyond the target dataset. In practice, the maximum channel number ($K_{max}$) was set to 20-40, with a maximum sequence length of 198, unless otherwise specified. All experiments were conducted on NVIDIA A100 GPUs with 40G GPU memory. Importantly, our zero-shot foundation model was trained on \textbf{Chronos dataset} without exposure to any data from the evaluated downstream tasks or datasets. The results of Section~\ref{zeroshot} are derived from a single pre-trained checkpoint, evaluated with or without fine-tuning based on the settings. The only exception is the long-term zero-shot experiments, which require extended sequence inputs while still utilizing the same pre-training dataset. To facilitate reproducibility and further research, we will release the pre-trained checkpoint.

\paragraph{Inference}  In the finetuning and inference stage, the choice of mask is tailored to align with the specific requirements of the user. This flexibility allows \modelnamespace to apply the most appropriate masking strategy based on the context of the task and application.  During inference, while the mask type and parameters are fixed for a given task to ensure consistency, TimeDiT's generative task architecture allows for flexible transformation of various downstream tasks. This adaptability enables us to address a wide range of time series challenges within a unified framework. Let $n$ represent the number of samples generated for each prediction, which we set to $n=10$ ($n=30$ for forecasting tasks) in our experimental setup at inference time. 
\begin{wraptable}{l}{0.5\textwidth}
\centering
\caption{{Comparison of inference times for single-sample generation.}}
\begin{tabular}{lc}
\toprule
Model & Inference Time (mm:ss) \\
\hline
Diffusion-TS & 00:06 \\
CSDI & 00:02 \\
TimeDiT & 00:01 \\
\bottomrule
\end{tabular}
\end{wraptable}
We use the median of these $n$ predictions as the final prediction, providing the added benefit of obtaining a confidence interval for \modelname's predictions. To prevent channel padding from affecting the generated samples, we mask out the invalid channels during sampling at each diffusion timestep so that \modelnamespace does not falsely treat the information in the non-valid channels as meaningful information. Padding is applied at the beginning of the temporal dimension to ensure that the most relevant information remains at the end, thereby mitigating the effect of padding. {We have included inference time comparisons for single-sample generation, where TimeDiT demonstrates superior computational efficiency, requiring only 1 second for single-sample generation, making it more practical for real-world applications. }

\begin{table}[htbp]
\centering
\caption{{A comparable analysis of representative general purposes time series models}}
\centering
\resizebox{\textwidth}{!}{
\begin{tabular}{lcccccc}
\toprule
\textbf{Model} & \textbf{Parameter Size} & \textbf{Model Architecture} & \textbf{Channel Setting} & \textbf{Task Type} & \textbf{Pretrain Dataset} & \textbf{Data Size} \\
\hline
Lag-LLama & - & Transformer & Univariate & Forecasting & Monash~\citep{godahewa2monash} & 300 Million Time Points \\
\hline
Moriai & \begin{tabular}[c]{@{}l@{}}S: 14M\\ B: 91M\\ L: 311M\end{tabular} & Transformer & Univariate & Forecasting & LOTSA~\citep{woounified} & 27 Billion Time Points \\
\hline
TimeDiT & \begin{tabular}[c]{@{}l@{}}S: 33M\\ B: 130M\\ L: 460M\\ XL: 680M\end{tabular} & Transformer + Diffusion & Multivariate & \begin{tabular}[c]{@{}l@{}}Forecasting,\\ Imputation,\\ Anomaly Detection,\\ Data Generation\end{tabular} & Chronos~\cite{ansari2024chronos} & About 5 Billion Time Points \\
\bottomrule
\end{tabular}
}
\label{table:7_com}
\end{table}

\section{Preliminaries of Diffusion Models}
In recent years, diffusion models have emerged as a promising approach to generative modeling. A diffusion process is a Markov chain that incrementally adds Gaussian noise to data over a sequence of steps, effectively destroying the data structure in the forward process and reconstructing the data structure during the reverse process.

\textbf{The forward process} adds noise to the data \( \mathbf{x}_0 \) over a series of timesteps \( t \) according to a variance schedule \( \beta_t \), resulting in a set of noisy intermediate variables \( \mathbf{x}_1, \mathbf{x}_2, \ldots, \mathbf{x}_T \). Each subsequent \( \mathbf{x}_t \) is derived from the previous step by applying Gaussian noise:
\begin{equation}
    q(\mathbf{x}_t \mid \mathbf{x}_{t-1}) = \mathcal{N}(\mathbf{x}_t; \sqrt{1 - \beta_t} \mathbf{x}_{t-1}, \beta_t \mathbf{I})
    \label{fwd}
\end{equation}
\textbf{The reverse process} aims to denoise the noisy variables step by step, sampling each \( \mathbf{x}_{t-1} \) from the learned distribution \( p_\theta(\mathbf{x}_{t-1} \mid \mathbf{x}_t) \). This distribution, modeled by a neural network parameterized by \( \theta \), approximates the Gaussian distribution:
\begin{equation}
p_\theta(\mathbf{x}_{t-1} \mid \mathbf{x}_t) = \mathcal{N}(\mathbf{x}_{t-1}; \mu_\theta(\mathbf{x}_t, t), \Sigma_\theta(\mathbf{x}_t, t))
\label{bwd}
\end{equation}
By iterating this reverse process from $t=T$ down to $t=0$, the model gradually reconstructs the original data from noise. Learning to clean \(\mathbf{x}_T\) through the reversed diffusion process reduces to building a surrogate approximator to parameterize \(\mu_\theta(\mathbf{x}_t, t)\) for all \(t\). The reverse process learns to predict the mean and covariance of each intermediate distribution, effectively approximating the original data distribution.

\section{Experiments setting}
\subsection{Datasets}

Please refer to  \textbf{Chronos }
(\url{https://huggingface.co/datasets/autogluon/chronos_datasets}) for our pre-trained dataset and the evaluation dataset are listed as follows:

\begin{enumerate}
    \item The ETT (Electricity Transformer Temperature) datasets 
 \citep{zhou2021informer}\footnote{ETT: \url{https://github.com/zhouhaoyi/ETDataset}} include electricity load data at various resolutions (ETTh \& ETTm) from two different electricity stations.
    \item  The Weather dataset \citep{zhou2021informer}\footnote{Weather:\url{https://www.ncei.noaa.gov/
data/local-climatological-data/}} comprises 21 meteorological indicators collected in Germany over the span of one year.
    \item The Electricity (ECL, Electricity Consuming Load) \citep{zhou2021informer}\footnote{ECL: \url{https://archive.ics.uci.edu/ml/
datasets/ElectricityLoadDiagrams20112014}} dataset provides information on electricity consumption.
    \item The SMD dataset \citep{SMD} includes multivariate time-series data collected from server machines in a data center. It typically contains metrics such as CPU usage, memory usage, and disk activity.
    \item The PSM dataset \citep{psm} is used for predictive maintenance and includes sensor data from industrial machines. It often contains readings such as temperature, pressure, and vibration over time.
    \item The MSL dataset \citep{MSL_SMAP} comes from the Mars Science Laboratory mission, specifically the Curiosity rover. It includes telemetry data from the rover's sensors and systems.
    \item The SWaT dataset \citep{SWaT} originates from a scaled-down water treatment testbed designed to reflect a real-world water treatment process. It includes sensor and actuator data collected over time.
    \item The SMAP dataset \citep{MSL_SMAP} comes from NASA's Soil Moisture Active Passive (SMAP) mission, which measures soil moisture and freeze/thaw state. It includes time-series data from multiple sensors aboard the SMAP satellite.
    \item The Sine dataset \citep{yoon2019time} is synthetically generated by sinusoidal waves.
    \item The Air Quality dataset~\citep{10.5555/3060832.3060999} \footnote{Air Quality: \url{https://archive.ics.uci.edu/dataset/360/air+quality}}contains hourly averaged readings from five metal oxide chemical sensors integrated into an Air Quality Chemical Multisensor Device. This device was positioned at road level in a highly polluted area of an Italian city. Data were collected from March 2004 to February 2005, making it the longest freely available record of on-field air quality chemical sensor responses.
    \item The Stock dataset \citep{yoon2019time}\footnote{Stock: \url{https://finance.yahoo.com/quote/GOOG}} contains daily historical Google stocks data from 2004 to 2019. 
    \item The UCI Appliances Energy prediction dataset \citep{yoon2019time}\footnote{Energy: \url{https://archive.ics.uci.edu/ml/datasets}}consists of multivariate, continuous-valued measurements including numerous temporal features measured at close intervals.
    \item The Weather\_2 dataset~\citep{godahewa2021monash}: The Weather\_2 dataset comprises hourly climate TSD collected near Monash University, Clayton, Victoria, Australia, from January 2010 to May 2021. It includes series for temperature, dewpoint temperature, wind speed, mean sea level pressure, relative humidity, surface solar radiation, surface thermal radiation, and total cloud cover.

    \item 
The PhysioNet dataset~\citep{silva2012predicting}\footnote{The PhysioNet: \url{https://physionet.org/content/challenge-2012/1.0.0/}} contains clinical time series data from 12,000 ICU patients, each with 42 vital variables recorded over 48 hours with naturally missing values. Patients are evenly divided into three groups of 4,000 each. For benchmarking purposes, we select 7 out of 42 variables. To address varying scales, we apply standard normalization, resulting in features with zero mean and unit variance.

    \item MIMIC-III~\citep{bica2020time}\footnote{MIMIC-III: \url{MIMIC-III: https://physionet.org/content/mimiciii/1.4/}}: MIMIC-III dataset contains 5000 patient ICU records with 19 variables from the lab events table including `anion gap, albumin, bands, bicarbonate, bilirubin, creatinine, chloride, glucose, hematocrit, hemoglobin, lactate, platelet, potassium, PTT, INR, PT, sodium, BUN, WBC'. They are irregularly sampled and we process them following the previous works ~\citep{bica2020time, cao2023estimating}, which have naturally missing values.
    
    \item NASDAQ: NASDAQ Top 10 Stocks dataset comprises time series data for the ten largest companies by market capitalization listed on the NASDAQ stock exchange. The dataset includes daily and 5-day price data for each stock from 2014-2024, offering two temporal resolutions for comprehensive analysis. We predict the close prices of each company in the multi-resolution forecasting task.

    \item Monash dataset archive~\citep{godahewa2monash}: The Monash repository contains 30 datasets, including publicly available time series datasets in various formats and those curated by us. Many datasets have different versions based on frequency and the inclusion of missing values. We use their multivariate time series version for pre-training and evaluation (specified if needed).
    
\end{enumerate}

\begin{table}[ht]
\caption{Dataset details}
\label{tab:dataset}
\begin{center}
\begin{small}
\begin{tabular}{l|cccc}
\toprule
Dataset & Domain & Length & Dimension & Frequency \\
\midrule
ETTh & Energy & 17420 & 7 & 1 hour\\
ETTm & Energy &69680 & 7 & 15 min\\
Weather & Nature & 52696 & 21 & 10 min \\
Electricity & Energy & 26304 & 321 & 1 hour \\
Air Quality & Nature & 9357 & 13 & 1 hour \\
Sine & Synthetic & 10000 & 5 & N/A \\
Stock & Finance & 3685 & 6 & 1 day \\
Energy & Energy & 19745& 28 & 10 min \\
MSL & Space& 132046  & 55& 1 min\\
PSM & Cloud &220322& 25& 1 min \\
SMAP & Space & 562800 & 25& 1 min\\
SMD & Cloud &1416825& 38 & 1 min\\
SWaT & Energy & 944920 & 51 & 1 second\\
Requests Minute &Cloud & 64800 & 10 & 1 min \\
Function Delay Minute &Cloud & 64800 & 10 & 1 min \\
Platform Delay Minute &Cloud & 64800 & 10 & 1 min \\
Memory Usage Minute &Cloud & 64800 & 10 & 1 min \\
CPU Limit Minute &Cloud & 64800 & 10 & 1 min \\
Memory Limit Minute &Cloud & 64800 & 10 & 1 min \\
Instances Minute &Cloud & 64800 & 10 & 1 min \\
Weather\_2 & Climate & 3001 & 695 & 1 day\\
PEMS\_SF & Traffic & 4320 & 852 & 1 hour\\
PhysioNet(b) & Health Care & - & 7 & Irregular \\
PhysioNet(b) & Health Care & - & 7 & Irregular \\
PhysioNet(c) & Health Care & - & 7 & Irregular \\
MIMIC-III & Health Care & - & 19 & 1 day \\
NASDAQ & Finance & 2516 & 20 & Multiresolution \\
\bottomrule
\end{tabular}
\end{small}
\end{center}
\vskip -0.1in
\end{table}

\subsection{Metrics}

\paragraph{MAE} describes the mean absolute error that measures the absolute difference between ground truth and prediction. 
\begin{equation}
        \text{MAE} = \frac{1}{n} \sum_{i=1}^{n} |y_i - \hat{y}_i|
    \end{equation}
\paragraph{MSE} describes the mean squared difference between ground truth and prediction. \begin{equation}
        \text{MSE} = \frac{1}{n} \sum_{i=1}^{n} (y_i - \hat{y}_i)^2
    \end{equation}
\paragraph{RMSE} is the sqaure root of MSE. 
\begin{equation}
        \text{RMSE} = \sqrt{\frac{1}{n} \sum_{i=1}^{n} (y_i - \hat{y}_i)^2}
    \end{equation}
\paragraph{Discriminative score} Following TimeGAN, we train a post-hoc time-series classification model (by optimizing a 2-layer LSTM) to distinguish between sequences from the original and generated datasets. First, each original sequence is labeled real, and each generated sequence is labeled not real. Then, an off-the-shelf (RNN) classifier is trained to distinguish between the two classes as a standard supervised task. We then report the classification error on the held-out test set.
\paragraph{Predictive Score} Following TimeGAN, we train a post-hoc sequence-prediction model (by optimizing a 2-layer LSTM) to predict next-step temporal vectors over each input sequence. Then, we evaluate the trained model on the original dataset. Performance is measured in terms of the mean absolute error (MAE); for event-based data, the MAE is computed as the absolute value of 1 - estimated probability that the event occured. 
\paragraph{Computations of CRPS}

We explain the definition and calculation of the CRPS metric. The continuous ranked probability score (CRPS)  assesses how well an estimated probability distribution \(F\) aligns with an observation \(x\). It is defined as the integral of the quantile loss \(\Lambda_{\alpha}(q, z) := (\alpha - \mathbf{1}_{z<q})(z - q)\) over all quantile levels \(\alpha \in [0, 1]\):

\begin{equation}
  \text{CRPS}(F^{-1}, x) = \int_{0}^{1} 2\Lambda_{\alpha}(F^{-1}(\alpha), x) \, d\alpha  
\end{equation}

where \(\mathbf{1}\) represents the indicator function. We then calculated quantile losses for quantile levels discretized in 0.05 increments. Thus, we approximated CRPS as follows:

\begin{equation}
\text{CRPS}(F^{-1}, x) \approx \frac{1}{19} \sum_{i=1}^{19} 2\Lambda_{i \cdot 0.05}(F^{-1}(i \cdot 0.05), x).
\end{equation}

Next, we computed the normalized average CRPS for all features and time steps:

\begin{equation}
\text{CRPS Score} = \frac{\sum_{k,l} \text{CRPS}(F^{-1}_{k,l}, x_{k,l})}{\sum_{k,l} |x_{k,l}|}
\end{equation}

where \(k\) and \(l\) denote the features and time steps of the imputation targets, respectively. The lower the CRPS, the more accurate the model, i.e., the closer the predicted probability is to the observed outcome.

\paragraph{Computations of CRPS\_{sum}} CRPS\_{sum} measures CRPS for the distribution \(F\) of the sum of all \(K\) features, calculated by:

\begin{equation}
\text{CRPS\_sum Score} = \frac{\sum_{l} \text{CRPS}(F^{-1}, \sum_{k} x_{k,l})}{\sum_{k,l} |x_{k,l}|}
\end{equation}

where \(\sum_{k} x_{k,l}\) is the total of the forecasting targets for all features at time point \(l\).

\paragraph{Precision} 
Precision measures the accuracy of positive predictions made by a model. It is defined as the ratio of true positives (TP) to the total number of predicted positives, which includes both true positives and false positives (FP). Mathematically, precision is expressed as:

\begin{equation}
    \text{Precision} = \frac{TP}{TP+FP}
\end{equation}

\paragraph{Recall} Recall, also known as sensitivity, measures a model's ability to correctly identify true positive instances. It is calculated as the ratio of true positives (TP) to the sum of true positives and false negatives (FN). In the context of anomaly detection, failing to detect an anomalous timestamp can have serious consequences, making recall a critical metric. Mathematically, recall is defined as:
\begin{equation}
    \text{Recall} = \frac{TP}{TP+FN}
\end{equation}

\paragraph{F1-score}
The F1-score is a balanced measure of model performance that combines Recall and Precision. It is calculated as the harmonic mean of these two metrics, giving equal importance to both. This score effectively captures the trade-off between Recall and Precision, penalizing significant disparities between them. By providing a single, comprehensive metric, the F1-score offers a more holistic view of a model's effectiveness, particularly useful when dealing with imbalanced datasets.
\begin{equation}
    \text{F1} = 2\times\frac{Precision \times Recall}{Precision + Recall}
\end{equation}
\subsection{Baselines}

We conduct a comprehensive comparative analysis, benchmarking TimeDiT against a diverse array of leading models in the field.
Our analysis extends to state-of-the-art probabilistic models, encompassing TimeGAN \citep{yoon2019time}, TimeVAE \citep{desai2021timevae}, Diffusion-TS \citep{yuan2024diffusion}, CSDI \citep{tashiro2021csdi}, TimeGrad \citep{rasul2021autoregressive}, TransMAF \citep{rasul2020multivariate}, GP-copula \citep{salinas2019high}, and TSDiff \citep{kollovieh2023predict}. We also evaluate against cutting-edge deterministic models, including DLinear \citep{zeng2023transformers}, GPT-2 \citep{one_fits_all}, TimesNet \citep{timesnet}, PatchTST \citep{Yuqietal-2023-PatchTST}, ETSformer \citep{woo2022etsformer}, FEDformer \citep{zhou2022fedformer}, LightTS \citep{lightts}, Autoformer \citep{wu2021autoformer}, and Anomaly Transformer \citep{xu2021anomaly}, LatentODE and LatentCDE\citep{rubanova2019latent}, etc.
Furthermore, we include comparisons with recent forecasting foundation models, such as TEMPO \citep{cao2023tempo}, Moirai \citep{woounified}, and LagLLama \citep{rasul2023lag}. This extensive comparison allows us to thoroughly evaluate TimeDiT's performance across a wide spectrum of methodologies and architectures in time series modeling.

\subsection{Physics Equations in Physics-Informed TimeDiT}
The Burgers Equation is:
\begin{equation}
    \frac{\partial u}{\partial t} + u\frac{\partial u}{\partial x} - v\frac{\partial^2 u}{\partial x^2} = 0
\end{equation}
where $v$ is the diffusion term. We set the $v$ (diffusion term) as 0.1 and randomly sample a combination of sine waves as initial status

The Advection Equation is:
\begin{equation}
    \frac{\partial u}{\partial t} + c\frac{\partial u}{\partial x} = 0
\end{equation}
where $c$ is the advection speed. We set the $c$ as 1.0 and randomly placed Gaussian peaks as initial status

The diffusion-reaction Equation is:
\begin{equation}
    \frac{\partial u}{\partial t} - D\frac{\partial^2 u}{\partial x^2} -R(u)= 0
\end{equation}
where $D$ is the diffusion coefficient and $R(u)$ is the reaction term. Here, we apply a linear reaction term $R(u) = -k\cdot u$, where $k$ is the reaction speed. We set the $D$ as 1.0, $k$ as 0.1, and a Gaussian distribution with random parameters as initial status.

The Kolmogrov Flow is a specific case of NS equation. More specifically, it is described by:

\begin{equation}
    \mathbf{u}(x, y, z, t) = \left( -\frac{\partial \psi}{\partial y}, \frac{\partial \psi}{\partial x}, 0 \right)
\end{equation}

where the $psi$ is the flow function. It is usually set as:

\begin{equation}
    \psi(x, y, z, t) = A \sin(kx) \cos(zy + \omega t)
\end{equation}

where $A,k,w$ are hyperparameters.
\section{Further discussion on Physics-Informed TimeDiT}

\subsection{Proof of Physics-Informed TimeDiT Theorem~\ref{th1}}
\begin{proof}
Let us consider the objective function:
    \begin{equation}
        \begin{aligned}
            O(q(y|x)) &= \mathbb{E}_{y\sim q(y|x)}K(y) - \alpha D_{KL}(q(y|x)||p(y|x))\\
            &=\mathbb{E}_{y\sim q(y|x)}K(y) - \alpha \int_yq(y|x)\log(\frac{q(y|x)}{p(y|x)})dy\\
            &=\int_yq(y|x)[K(y) + \alpha \log p(y|x) -\alpha \log q(y|x)]dy\\
        \end{aligned}
    \end{equation}
    We try to find the optimal $q(y|x)$ through Lagrange multipliers. The constraint of the above objective function is that $q(y|x)$ is a valid $\int_y q(y|x)dy=1$. Thus, the Lagrangian is:
    \begin{equation}
        \begin{aligned}
            L(q(y|x),\lambda) &= \int_yq(y|x)[K(y) + \alpha \log p(y|x) -\alpha \log q(y|x)]dy - \lambda (\int_y q(y|x)dy-1)\\
            &=\int_y q(y|x)[K(y) + \alpha \log p(y|x) -\alpha \log q(y|x)-\lambda q(y|x)]dy + \lambda\\
        \end{aligned}
    \end{equation}
    We define $f(q(y|x),y,\lambda)=q(y|x)[K(y) + \alpha \log p(y|x) -\alpha \log q(y|x) - \lambda] + \lambda h(y)]$, where $h(y)$ can be the density function of any fixed distribution defined on the support set of $y$. Therefore, $L(q(y|x),\lambda) = \int_y f(q(y|x),y,\lambda)dy$.
    According to Euler-Lagrange equation, when the above Lagrangian achieve extreme point, we have:
    \begin{equation}
        \begin{aligned}
        \frac{\partial f}{\partial q} = K(y) + \alpha \log p(y|x) -\alpha \log q(y|x) - \lambda - \alpha = 0
        \end{aligned}
    \end{equation}
    Thus, we have:
    \begin{equation}
        \begin{aligned}
        \alpha\log q(y|x) &= K(y) + \alpha \log p(y|x) - \log q(y|x) - \lambda - \alpha\\
        q(y|x) &= \exp(\frac{1}{\alpha}K(y) + \log p(y|x) - \frac{\lambda}{\alpha} - 1)\\
        &=\frac{1}{\exp(\frac{\lambda}{\alpha} +1)}\exp(\frac{1}{\alpha}K(y) + \log p(y|x))\\
        \end{aligned}
    \end{equation}
    Meanwhile, since $\int_y q(y|x)dy=1$, we have:
    \begin{equation}
    \begin{aligned}
        \int_y \exp(\frac{1}{\alpha}K(y) + \log p(y|x) - \frac{\lambda}{\alpha} - 1)dy&=1\\
        \frac{1}{\exp(\frac{\lambda}{\alpha} +1)}\int_y \exp(\frac{1}{\alpha}K(y) + \log p(y|x))dy&=1\\
    \end{aligned}
    \end{equation}
    Thus, we have $\exp(\frac{\lambda}{\alpha} +1) = \int_y \exp(\frac{1}{\alpha}K(y) + \log p(y|x))dy=Z$, leading to:
    \begin{equation}
        q(y|x) = \frac{1}{Z}\exp(K(y)+\alpha\log p(y|x)), Z = \int \exp(K(y)+\alpha\log p(y|x))dy
    \end{equation}
\end{proof}

\subsection{{Physics-Informed TimeDiT vs. Direct PDE-based Generation Training}}

\begin{table}[ht]
\centering
\caption{{Comparison on the physics informed zero-shot TimeDiT with fully trained baselines.}}
\resizebox{0.6\textwidth}{!}{
\begin{tabular}{llcccc}
\toprule
\multicolumn{6}{c}{\textbf{Burgers}} \\
\hline
&& MSE & RMSE & MAE & CRPS \\
\hline
\multirow{3}{*}{Full-shot}&DLinear & 0.031(0.002) & 0.175(0.001) & 0.12610.005) & 1.400(0.057) \\
&PatchTST & 0.029(0.001) & 0.170(0.001) & 0.125(0.004) & 1.411(0.051) \\
&NeuralCDE & 0.031(0.002) & 0.176(0.002) & 0.126(0.005) & 1.397(0.061) \\\hline
Zero-shot &TimeDiT & \textbf{0.011(0.001)} & \textbf{0.104(0.005)} & \textbf{0.083(0.003)} & \textbf{1.395(0.053)} \\
\hline
\hline
\multicolumn{6}{c}{\textbf{Vorticity}} \\
\hline
&& MSE & RMSE & MAE & CRPS \\
\hline
\multirow{3}{*}{Full-shot}&DLinear & 2.650(0.003) & 1.628(0.001) & 1.459(0.010) & 0.695(0.005) \\
&PatchTST & 2.651(0.002) & 1.628(0.002) & 1.460(0.012) & 0.700(0.001) \\
&NeuralCDE & 2.631(0.001) & 1.622(0.001) & 1.453(0.010) & 0.691(0.005) \\\hline
Zero-shot &TimeDiT & \textbf{1.524(0.523)} & \textbf{1.234(0.021)} & \textbf{0.772(0.009)} & \textbf{0.445(0.006)} \\
\bottomrule
\end{tabular}}
\end{table}

{
The tension between physical constraints and learned distributions in TimeDiT is managed through a sophisticated energy-based optimization framework that combines two key components:}
\begin{itemize}
    \item {the physics knowledge represented by function $K(x^{\text{tar}}; F)$, which measures PDE residuals for physical law conformity}
    \item {the learned probabilistic distribution $p(x^{\text{tar}} | x^{\text{con}})$ from the diffusion model}
\end{itemize}

{This balance is achieved through an energy function:
\[
E(x^{\text{tar}}; x^{\text{con}}) = K(x^{\text{tar}}; F) + \alpha \log p(x^{\text{tar}} | x^{\text{con}})
\]
where the parameter $\alpha$ controls the trade-off between physical consistency and distribution fidelity.}

{Rather than directly modifying model parameters, TimeDiT implements this balance through an iterative sampling procedure that:}
\begin{enumerate}
    \item {starts with samples from the learned distribution}
    \item {gradually refines them using physical gradients while maintaining probabilistic characteristics}
\end{enumerate}

{This approach allows the model to generate samples that respect both the learned patterns in the data and the underlying physical laws without significantly compromising either aspect, ultimately resolving the tension through a theoretically-grounded Boltzmann distribution as the optimal solution.}

{Physics-informed machine learning represents an active research area where physical constraints guide model outputs toward realistic solutions~\citep{meng2022physics}. Our physics-informed TimeDiT offers a novel approach that addresses key limitations of traditional PDE-based training methods. While direct use of PDE-based solvers to generate samples and then training is possible, TimeDiT provides crucial advantages in efficiency and flexibility. Our model incorporates physical knowledge during inference through energy-based sampling that guides the reverse diffusion process. This means we can flexibly integrate different physical constraints without any model retraining or parameter updates. We conduct an experimental comparison with direct PDE-based training methods. Using PDE solvers, we generated 5,000 training samples per scenario and trained three baseline models: DLinear, PatchTST, and NeuralCDE. Notably, zero-shot TimeDiT outperformed these models. For 6 PDE equations, the traditional approach required training 18 distinct models, resulting in significant computational overhead - approximately 18 times more training time - and extensive code modifications. This approach becomes increasingly impractical in real-world applications where multiple physical laws interact, as each new constraint would require training additional dedicated models. In contrast, TimeDiT's unified framework incorporates various physical constraints during inference while maintaining a single trained model, providing a more efficient and scalable solution.}

\section{Detailed Experiment Results}

\subsection{Forecasting}

\begin{figure}[htbp]
\centering
\subfigure[Miss Value in Exchange dataset]{
\begin{minipage}{0.5\linewidth}
\centering
\includegraphics[width=2.6in]{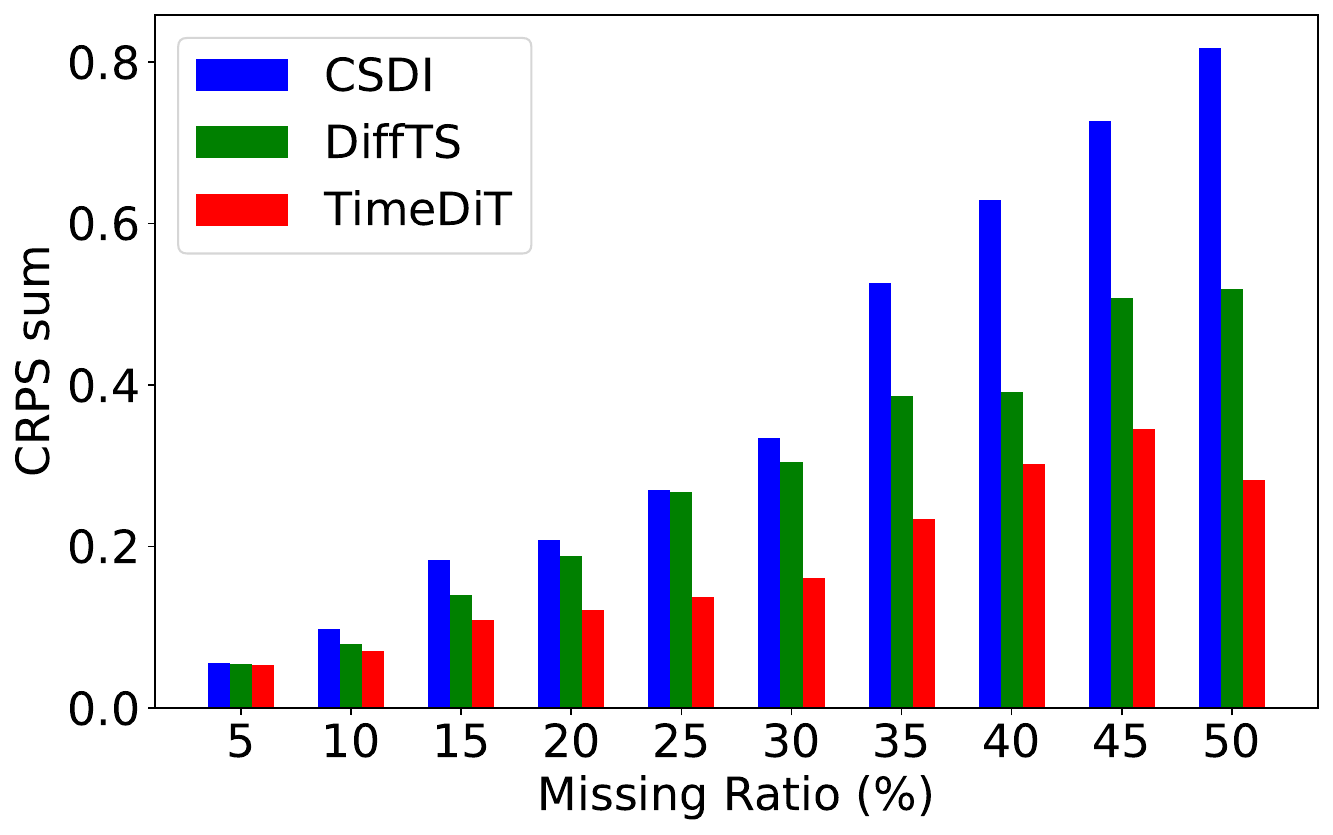}
\end{minipage}%
}%
\subfigure[Multi-resolution in Exchange dataset]{
\begin{minipage}{0.5\linewidth}
\centering
\includegraphics[width=2.6in]{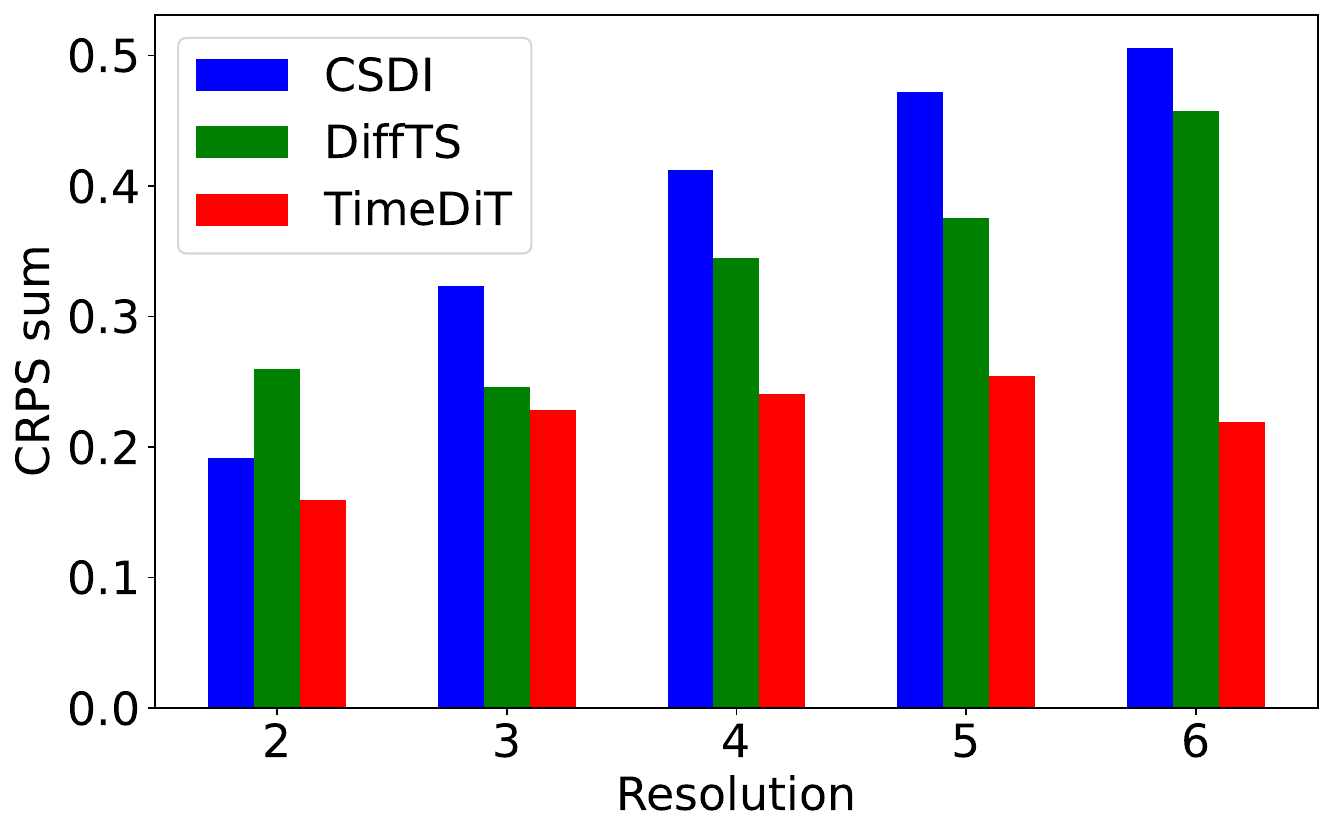}
\end{minipage}%
}

\caption{Visualization of miss value (a) and multiresolution (b) forecasting results on the Exchange dataset and miss value (c) and multiresolution (d) forecasting results on the Traffic dataset. Compared between our model \modelnamespace and state-of-the-art diffusion-based methods. The x-axis number in (b) is the sampling skip in the resolutions in the multivariate input. }
\label{fore_fig}
\end{figure}

\begin{figure}[htbp]
\centering
\subfigure[Miss Value in Traffic dataset]{
\begin{minipage}{0.5\linewidth}
\centering
\includegraphics[width=2.6in]{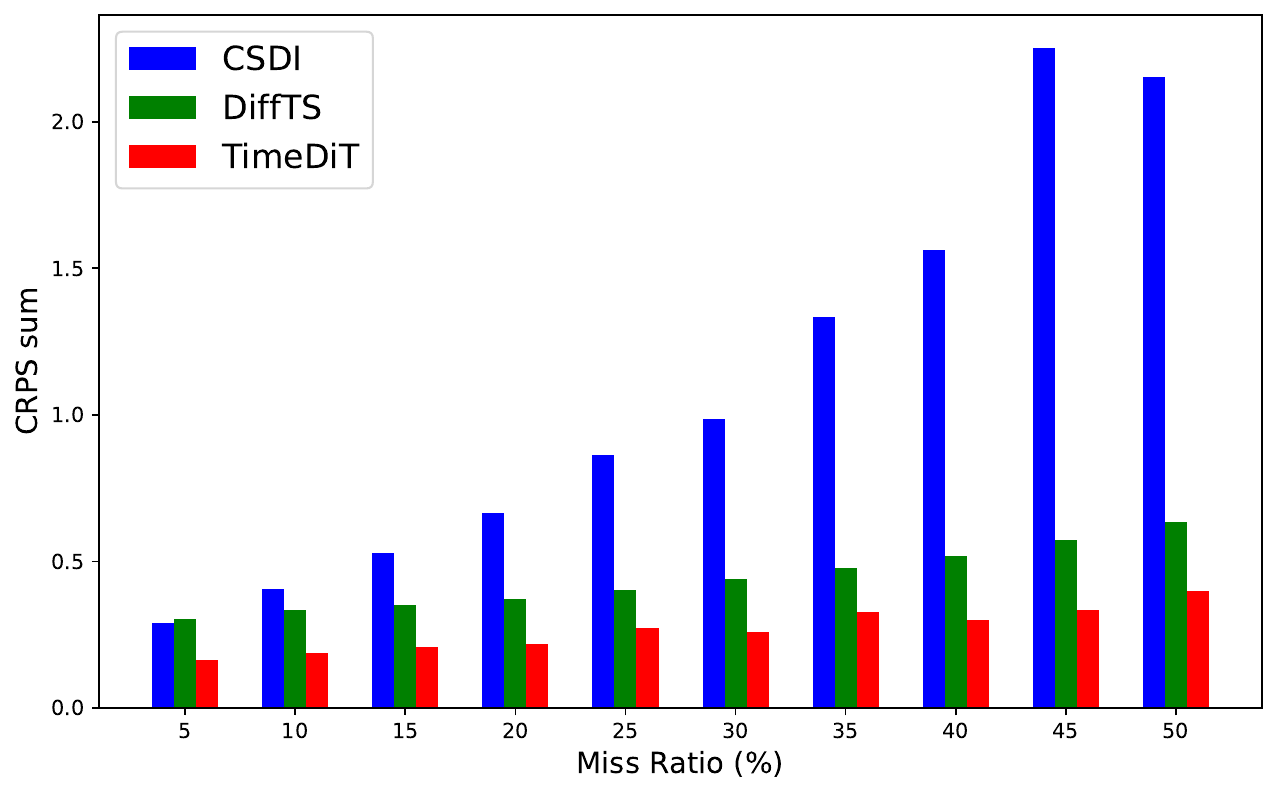}
\end{minipage}%
}%
\subfigure[Multi-resolution in Traffic dataset]{
\begin{minipage}{0.5\linewidth}
\centering
\includegraphics[width=2.6in]{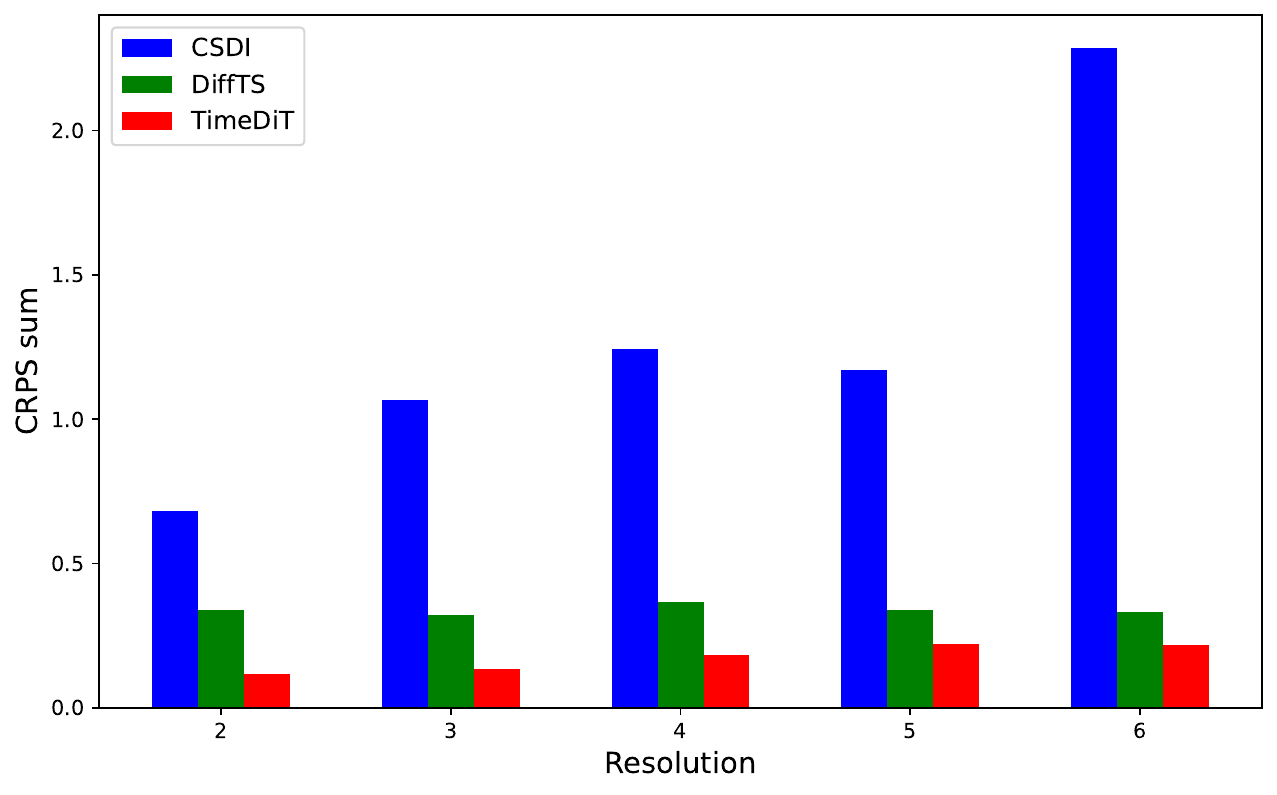}
\end{minipage}%
}%
\caption{Visualization of miss value (a) and multi resolution (b) forecasting results on the Traffic (PEMS-SF) dataset. Compared between our model \modelnamespace and state-of-the-art diffusion-based methods. The x-axis number in (b) is the sampling skip in the resolutions in the multivariate input. }
\label{fore_fig_2}
\end{figure}

\subsubsection{Practical Forecasting Setting}
\paragraph{Setting of Table~\ref{tab:1_MS_RES}.} The Nasdaq dataset features two resolutions (daily and 5-day intervals), using 168 historical steps to predict 30 future steps. The Air Quality dataset, containing natural missing values, also uses 168 steps to predict 30. For healthcare datasets, we group and normalize patient records individually. In PhysioNet, we select trajectories longer than 10 steps, using 96 to predict 24. For MIMIC-III, we choose trajectories between 10 and 40 steps, using 27 to predict 3 due to shorter lengths. This diverse dataset collection enables comprehensive evaluation of TimeDiT across various temporal resolutions and domain-specific challenges, spanning financial forecasting, environmental monitoring, and healthcare predictive modeling. We compare TimeDiT with state-of-the-art models in two categories: deterministic forecasting models adapted with a Student's t-distribution head for probabilistic outputs, and inherently diffusion-based 
 probabilistic time series forecasting SOTA models. All baseline models are trained in a full-shot setting, while TimeDiT leverages a pre-trained foundation model, fine-tuning it on realistic datasets. Notably, TimeDiT can naturally handle input data with missing values, eliminating the need for additional imputation methods. This capability allows TimeDiT to perform forecasting directly using learned representations, even in the presence of incomplete data.
 
\subsubsection{More Practical Forecasting Results}
\paragraph{More results on miss-value and multi-resolution setting.}To further evaluate the practical ability of our proposed TimeDiT, we built two cases based on the previous dataset: the missing value scenario, where we created datasets with various missing ratios, simulating incomplete data often encountered in practice. In the multi-resolution setting, we sampled each individual time series within the multivariate dataset at different resolutions, reflecting the diverse sampling frequencies often present in real-world data collection.
Figure ~\ref{fore_fig}  and Figure ~\ref{fore_fig_2} illustrate TimeDiT's performance in realistic scenarios, showcasing its effectiveness across different sampling frequencies on the Exchange dataset. In Figure ~\ref{fore_fig}(a) and Figure ~\ref{fore_fig_2}(a), we observe TimeDiT's superior performance in handling missing data. As the missing ratio increases from 5\% to 50\%, TimeDiT maintains the lowest CRPS\_{sum} across all scenarios, indicating its robustness to data gaps. The performance gap between TimeDiT and other models widens as the missing ratio increases, highlighting its effectiveness in more challenging conditions.
Figure ~\ref{fore_fig}(b) and Figure ~\ref{fore_fig_2}(b) demonstrate TimeDiT's ability to manage multi-resolution data, where it maintains a clear performance advantage as the number of different sampling resolutions increases from 2 to 6. This demonstrates its ability to effectively integrate and forecast TSD sampled at varying frequencies.

\begin{table}[ht]
\caption{Evaluate time series dataset for forecasting tasks.}
\centering
\resizebox{0.8\textwidth}{!}{
\begin{tabular}{l|llp{2cm}p{2cm}p{2cm}ll}
\toprule
            & Features & Time Steps & History Length ($L_1$) & Prediction Horizon ($L_2$) & Rolling Windows & Frequency                      & Domain \\ \hline
Solar       & 137      & 7009       & 168                    & 24                         & 7               & 1 hour                      & $\mathbb{R}^{+}$    \\ 
Electricity & 370      & 5833       & 168                    & 24                         & 7               & 1 hour & $\mathbb{R}^{+}$      \\ 
Traffic     & 963      & 4001       & 168                    & 24                         & 7               & 1 hour & (0, 1) \\ 
Taxi        & 1214     & 1488       & 48                     & 24                         & 56              & 30 mins                        & $\mathbb{N}$      \\ 
Exchange    & 8        & 6071       & 60                     & 30                         & 5               & 1 day                          & $\mathbb{R}^{+}$      \\ \bottomrule
\end{tabular}}
\label{tab:fore_data}
\end{table}

\subsubsection{Full-shot Forecasting Setting}

For the full-shot benchmarking forecasting and zero-shot forecasting task, we utilized five widely-used open datasets to evaluate probabilistic time series forecasting performance. These datasets were collected in GluonTS ~\citep{gluonTS} and have been previously employed in ~\citep{tashiro2021csdi, salinas2019high}:

\begin{itemize}
    \item Solar\footnote{Solar: \url{https://www.nrel.gov/grid/solar-power-data.html}}: Hourly solar power production records from 137 stations in Alabama State, as used in ~\citep{lai2018modeling}.
   \item Electricity\footnote{Electricity:\url{https://archive.ics.uci.edu/ml/datasets/ElectricityLoadDiagrams20112014}}: Hourly time series of electricity consumption for 370 customers, as used in ~\citep{asuncion2007uci}.
   \item Traffic\footnote{Traffic\_nips: \url{https://archive.ics.uci.edu/dataset/204/pems\_sf}}: Hourly occupancy rates of 963 San Francisco freeway car lanes, with values between 0 and 1 ~\citep{asuncion2007uci}.
   \item Taxi\footnote{Taxi: \url{https://www1.nyc.gov/site/tlc/about/tlc-trip-record-data}}: Half-hourly spatio-temporal time series of New York taxi rides taken at 1,214 locations, using data from January 2015 for training and January 2016 for testing, as proposed in ~\citep{tlc2017nyc}.
   \item Exchange rate\footnote{Exchange: \url{https://github.com/laiguokun/multivariate-time-series-data}}: Daily exchange rates between 8 currencies, namely Australia,
the United Kingdom, Canada, Switzerland, China, Japan, New Zealand, and Singapore, as used in  ~\citep{lai2018modeling}.
\end{itemize}

Table ~\ref{tab:fore_data} summarizes the characteristics of each dataset. The task for these datasets is to predict the future $L_2$ steps given the observed $L_1$ steps. We set $L_1$ and $L_2$ values based on previous studies ~\citep{tashiro2021csdi, salinas2019high}. For training, we randomly selected $L_1 + L_2$ consecutive time steps as a single time series and designated the last $L_2$ steps as forecasting targets. We adhered to the train/test splits used in previous studies and utilized the last five samples of the training data as validation data.
For the full-shot setting, we trained separate models on different datasets. Due to the large number of features in multivariate time series, we adopted subset sampling of features for training. For each input, we split them into subsets based on their order. If the last subset was smaller than the fixed shape, we applied padding to ensure equal input sizes across all subsets.

\begin{table}[!ht]

\label{tab:full_impute}
\begin{center}
\caption{Full result of imputation task. }
\resizebox{\textwidth}{!}{
\setlength\tabcolsep{3pt}
\begin{tabular}{cc|cc|cc|cc|cc|cc|cc|cc|cc|cc|cc|cc|cc|cc|cc|cc|cc}
\toprule

\multicolumn{2}{c|}{Methods} &\multicolumn{2}{c|}{\modelname}
&\multicolumn{2}{c|}{{Timer}}
&\multicolumn{2}{c|}{{TimeMixer}}
&\multicolumn{2}{c|}{{iTransformer}}
&\multicolumn{2}{c|}{GPT2(3)} 
& \multicolumn{2}{c|}{TimesNet}
& \multicolumn{2}{c|}{PatchTST}
&\multicolumn{2}{c|}{ETSformer}
&\multicolumn{2}{c|}{LightTS}
&\multicolumn{2}{c|}{DLinear}
&\multicolumn{2}{c|}{FEDformer}
&\multicolumn{2}{c|}{Stationary}
&\multicolumn{2}{c|}{Autoformer}
&\multicolumn{2}{c}{Informer}
&\multicolumn{2}{c}{Reformer} \\
Mask&Ratio&MSE&MAE&MSE&MAE&MSE&MAE&MSE&MAE&MSE&MAE&MSE&MAE&MSE&MAE&MSE&MAE&MSE&MAE&MSE&MAE&MSE&MAE&MSE&MAE&MSE&MAE&MSE&MAE&MSE&MAE \\
\midrule
\multirow{5}{*}{\rotatebox{90}{$ETTh1$}}
& 12.5\% & \textbf{0.022} & \textbf{0.096}  &{0.119} &{0.222}&{0.094}&{0.203}&{0.099}&{0.221}
&\underline{0.043}&\underline{0.140}&0.057&0.159&0.093&0.201&0.126&0.263&0.240&0.345&0.151&0.267&0.070&0.190&0.060&0.165&0.074&0.182&0.114&0.234&0.074&0.194 \\
& 25\% & \textbf{0.029} & \textbf{0.110} &{0.133} &{0.235}&{0.111}&{0.219}&{0.125}&{0.249}
&\underline{0.054}&\underline{0.156}&0.069&0.178&0.107&0.217&0.169&0.304&0.265&0.364&0.180&0.292&0.106&0.236&0.080&0.189&0.090&0.203&0.140&0.262&0.102&0.227 \\
& 37.5\% & \textbf{0.039} & \textbf{0.127} &{0.151} &{0.249}&{0.124}&{0.233}&{0.158}&{0.281}
&\underline{0.072}&\underline{0.180}&0.084&0.196&0.120&0.230&0.220&0.347&0.296&0.382&0.215&0.318&0.124&0.258&0.102&0.212&0.109&0.222&0.174&0.293&0.135&0.261 \\
& 50\%  & \textbf{0.055} & \textbf{0.152} &{0.176} &{0.267}&{0.144}&{0.249}&{0.214}&{0.328}
&0.107&0.216&\underline{0.102}&\underline{0.215}&0.141&0.248&0.293&0.402&0.334&0.404&0.257&0.347&0.165&0.299&0.133&0.240&0.137&0.248&0.215&0.325&0.179&0.298 \\
& Avg & \textbf{0.036} & \textbf{0.122} &{0.145} &{0.243}&{0.119}&{0.226}&{0.149}&{0.270}
&\underline{0.069}&\underline{0.173}&0.078&0.187&0.115&0.224&0.202&0.329&0.284&0.373&0.201&0.306&0.117&0.246&0.094&0.201&0.103&0.214&0.161&0.279&0.122&0.245 \\
\midrule

\multirow{5}{*}{\rotatebox{90}{$ETTh2$}}
& 12.5\% & \textbf{0.019} & \textbf{0.083} &{0.070} &{0.163}&{0.056}&{0.145}&{0.099}&{0.221}
&\underline{0.039}&\underline{0.125}&0.040&0.130&0.057&0.152&0.187&0.319&0.101&0.231&0.100&0.216&0.095&0.212&0.042&0.133&0.044&0.138&0.305&0.431&0.163&0.289 \\
& 25\% & \textbf{0.024} & \textbf{0.098}&{0.074} &{0.168}&{0.063}&{0.157}&{0.130}&{0.254}
&\underline{0.044}&\underline{0.135}&0.046&0.141&0.061&0.158&0.279&0.390&0.115&0.246&0.127&0.247&0.137&0.258&0.049&0.147&0.050&0.149&0.322&0.444&0.206&0.331 \\
& 37.5\% & \textbf{0.032} & \textbf{0.116} &{0.079} &{0.174}&{0.064}&{0.158}&{0.158}&{0.281}
&0.051&\underline{0.147}&0.052&{0.151}&0.067&0.166&0.400&0.465&0.126&0.257&0.158&0.276&0.187&0.304&0.056&0.158&0.060&0.163&0.353&0.462&0.252&0.370 \\
& 50\% & \textbf{0.051} & \textbf{0.148} &{0.085} &{0.182}&{0.071}&{0.168}&{0.214}&{0.328}
&\underline{0.059}&\underline{0.158}&{0.060}&{0.162}&0.073&0.174&0.602&0.572&0.136&0.268&0.183&0.299&0.232&0.341&0.065&0.170&0.068&0.173&0.369&0.472&0.316&0.419 \\
& Avg & \textbf{0.031} & \textbf{0.111}   &{0.077} &{0.172}&{0.064}&{0.157}&{0.150}&{0.271}
&\underline{0.048}&\underline{0.141}&0.049&0.146&0.065&0.163&0.367&0.436&0.119&0.250&0.142&0.259&0.163&0.279&0.053&0.152&0.055&0.156&0.337&0.452&0.234&0.352 \\

\midrule
\multirow{5}{*}{\rotatebox{90}{$ETTm1$}}
& 12.5\% & \textbf{0.014} & \textbf{0.078} &{0.044} &{0.131}&{0.046}&{0.136}&{0.045}&{0.147}
&\underline{0.017}&\underline{0.085}&0.023&0.101&0.041&0.130&0.096&0.229&0.093&0.206&0.080&0.193&0.052&0.166&0.032&0.119&0.046&0.144&0.063&0.180&0.042&0.146 \\
& 25\% & \textbf{0.018} & \textbf{0.085} &{0.048} &{0.136}&{0.048}&{0.137}&{0.060}&{0.172}
&\underline{0.022}&\underline{0.096}&0.023&0.101&0.044&0.135&0.096&0.229&0.093&0.206&0.080&0.193&0.052&0.166&0.032&0.119&0.046&0.144&0.063&0.180&0.042&0.146 \\
& 37.5\% & \textbf{0.023} & \textbf{0.096} &{0.053} &{0.144}&{0.059}&{0.155}&{0.078}&{0.196}
&\underline{0.029}&\underline{0.111}&\underline{0.029}&\underline{0.111}&0.049&0.143&0.133&0.271&0.113&0.231&0.103&0.219&0.069&0.191&0.039&0.131&0.057&0.161&0.079&0.200&0.063&0.182 \\
& 50\% & \textbf{0.032} & \textbf{0.114} &{0.061} &{0.154}&{0.053}&{0.145}&{0.102}&{0.226}
&{0.040}&{0.128}&\underline{0.036}&\underline{0.124}&0.055&0.151&0.186&0.323&0.134&0.255&0.132&0.248&0.089&0.218&0.047&0.145&0.067&0.174&0.093&0.218&0.082&0.208 \\
& Avg & \textbf{0.022} & \textbf{0.093} &{0.051} &{0.141}&{0.051}&{0.143}&{0.071}&{0.185}
&0.028&\underline{0.105}&\underline{0.027}&0.107&0.047&0.140&0.120&0.253&0.104&0.218&0.093&0.206&0.062&0.177&0.036&0.126&0.051&0.150&0.071&0.188&0.055&0.166 \\
\midrule

\multirow{5}{*}{\rotatebox{90}{$ETTm2$}}
& 12.5\% & \underline{0.019} & \textbf{0.074} &{0.032} &{0.098}&{0.024}&{0.086}&{0.052}&{0.151}
&\textbf{0.017}&\underline{0.076}&{0.018}&0.080&0.026&0.094&0.108&0.239&0.034&0.127&0.062&0.166&0.056&0.159&0.021&0.088&0.023&0.092&0.133&0.270&0.108&0.228 \\
& 25\% & {0.029} & {0.096} &{0.034} &{0.102}&\underline{0.026}&{0.090}&{0.071}&{0.179}
&\textbf{0.020}&\textbf{0.080}&\textbf{0.020}&\underline{0.085}&0.028&0.099&0.164&0.294&0.042&0.143&0.085&0.196&0.080&0.195&{0.024}&0.096&0.026&0.101&0.135&0.272&0.136&0.262 \\
& 37.5\% & {0.039} & {0.114} &{0.036} &{0.106}&{0.029}&{0.094}&{0.091}&{0.204}
&\textbf{0.022}&\textbf{0.087}&\underline{0.023}&\underline{0.091}&0.030&0.104&0.237&0.356&0.051&0.159&0.106&0.222&0.110&0.231&0.027&0.103&0.030&0.108&0.155&0.293&0.175&0.300 \\
& 50\% & {0.050} & {0.132} &{0.040} &{0.112}&{0.032}&{0.101}&{0.117}&{0.232}
&\textbf{0.025}&\textbf{0.095}&\underline{0.026}& \underline{0.098}&0.034&0.110&0.323&0.421&0.059&0.174&0.131&0.247&0.156&0.276&0.030&0.108&0.035&0.119&0.200&0.333&0.211&0.329 \\
& Avg & {0.034} & {0.104} &{0.035} &{0.105}&{0.028}&{0.093}&{0.083}&{0.192}
&\textbf{0.021}&\textbf{0.084}&\underline{0.022}&\underline{0.088}&0.029&0.102&0.208&0.327&0.046&0.151&0.096&0.208&0.101&0.215&0.026&0.099&0.029&0.105&0.156&0.292&0.157&0.280 \\
\midrule

\multirow{5}{*}{\rotatebox{90}{$ECL$}}
& 12.5\% & \textbf{0.049} & \textbf{0.142}&{0.077} &{0.174}&{0.047}&{0.145}&{0.073}&{0.190}
&{0.080}&{0.194}&0.085&0.202&\underline{0.055}&\underline{0.160}&0.196&0.321&0.102&0.229&0.092&0.214&0.107&0.237&0.093&0.210&0.089&0.210&0.218&0.326&0.190&0.308 \\
& 25\% &  \textbf{0.057} & \textbf{0.153} &{0.087} &{0.184}&{0.055}&{0.156}&{0.090}&{0.214}
&{0.087}&{0.203}&0.089&0.206&\underline{0.065}&\underline{0.175}&0.207&0.332&0.121&0.252&0.118&0.247&0.120&0.251&0.097&0.214&0.096&0.220&0.219&0.326&0.197&0.312 \\
& 37.5\% & \textbf{0.067} & \textbf{0.175} &{0.101} &{0.199}&{0.064}&{0.169}&{0.107}&{0.234}
&0.094&{0.211}&{0.094}&0.213&\underline{0.076}&\underline{0.189}&0.219&0.344&0.141&0.273&0.144&0.276&0.136&0.266&0.102&0.220&0.104&0.229&0.222&0.328&0.203&0.315\\
& 50\% & \textbf{0.097} & \textbf{0.220} &{0.121} &{0.219}&{0.078}&{0.185}&{0.127}&{0.257}
&0.101&{0.220}&{0.100}&0.221&\underline{0.091}&\underline{0.208}&0.235&0.357&0.160&0.293&0.175&0.305&0.158&0.284&0.108&0.228&0.113&0.239&0.228&0.331&0.210&0.319 \\
& Avg & \textbf{0.068} & \textbf{0.172}  &{0.097} &{0.194}&{0.061}&{0.164}&{0.099}&{0.224}
&{0.090}&{0.207}&0.092&0.210&\underline{0.072}&\underline{0.183}&0.214&0.339&0.131&0.262&0.132&0.260&0.130&0.259&0.100&0.218&0.101&0.225&0.222&0.328&0.200&0.313 \\
\midrule

\multirow{5}{*}{\rotatebox{90}{$Weather$}}
& 12.5\% & {0.029} & \textbf{0.033} &{0.107} &{0.168}&{0.030}&{0.054}&{0.039}&{0.089}
&\underline{0.026}&0.049&\textbf{0.025}&\underline{0.045}&0.029&0.049&0.057&0.141&0.047&0.101&0.039&0.084&0.041&0.107&0.027&0.051&0.026&0.047&0.037&0.093&0.031&0.076 \\
& 25\% & {0.031} & \textbf{0.033} &{0.108} &{0.167}&{0.029}&{0.043}&{0.047}&{0.108}
&\textbf{0.028}&{0.052}&\underline{0.029}&{0.052}&0.031&0.053&0.065&0.155&{0.052}&0.111&0.048&0.103&0.064&0.163&0.029&0.056&0.030&0.054&0.042&0.100&0.035&0.082 \\
& 37.5\% & {0.034} & \textbf{0.037} &{0.108} &{0.167}&{0.032}&{0.047}&{0.055}&{0.121}
&\underline{0.033}&0.060&\textbf{0.031}&\underline{0.057}&0.035&0.058&0.081&0.180&0.058&0.121&{0.057}&0.117&0.107&0.229&0.033&0.062&\underline{0.032}&0.060&0.049&0.111&0.040&0.091 \\
& 50\% & \textbf{0.031} & \textbf{0.041} &{0.109} &{0.168}&{0.035}&{0.051}&{0.070}&{0.145}
&{0.037}&0.065&\underline{0.034}&\underline{0.062}&0.038&{0.063}&0.102&0.207&0.065&0.133&0.066&0.134&0.183&0.312&{0.037}&0.068&{0.037}&0.067&0.053&0.114&0.046&0.099 \\
& Avg & \underline{0.031} & \textbf{0.036} &{0.108} &{0.168}&{0.031}&{0.049}&{0.053}&{0.116}
&\underline{0.031}&0.056&\textbf{0.030}&\underline{0.054}&0.060&0.144&0.076&0.171&0.055&0.117&0.052&0.110&0.099&0.203&0.032&0.059&\underline{0.031}&0.057&0.045&0.104&0.038&0.087 \\
\bottomrule

\end{tabular}
}
\end{center}
\end{table}

\begin{figure}[htbp]
\centering
\subfigure[Imputation results on ETTh1 dataset.]{
\begin{minipage}{\linewidth}
\centering
\includegraphics[width=5in]{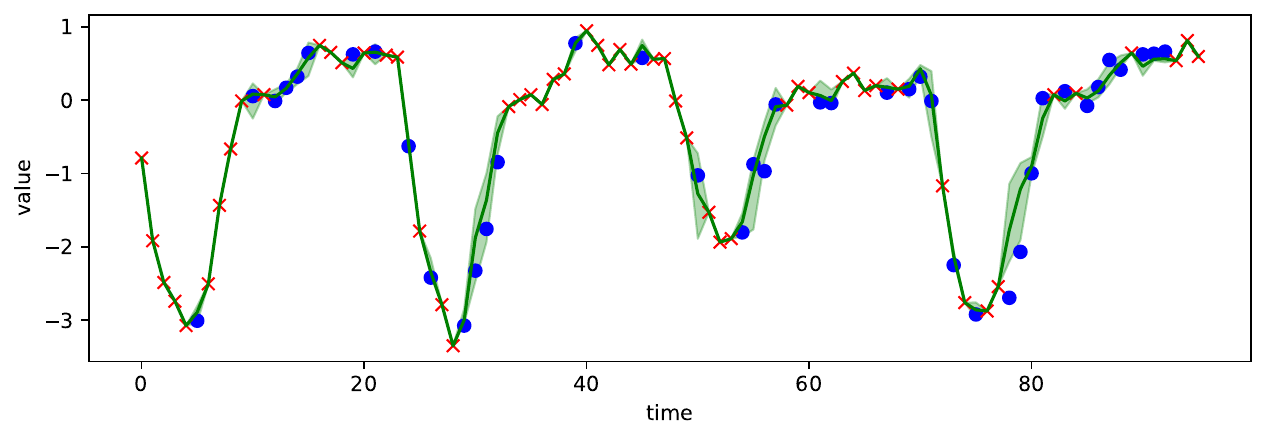}
\end{minipage}%
}%

\subfigure[Imputation results on ETTh2 dataset.]{
\begin{minipage}{\linewidth}
\centering
\includegraphics[width=5in]{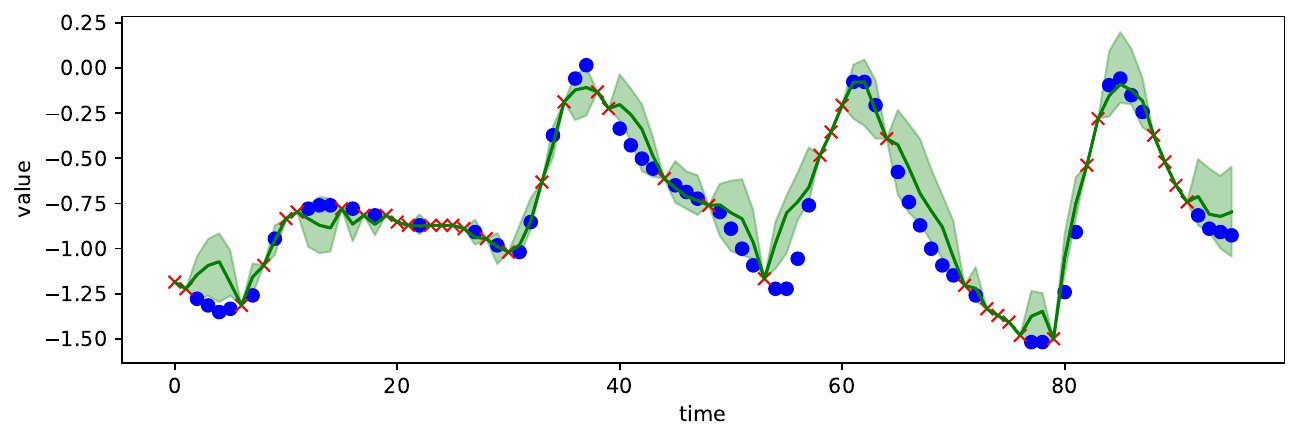}
\end{minipage}%
}%

\subfigure[Imputation results on ETTm1 dataset.]{
\begin{minipage}{\linewidth}
\centering
\includegraphics[width=5in]{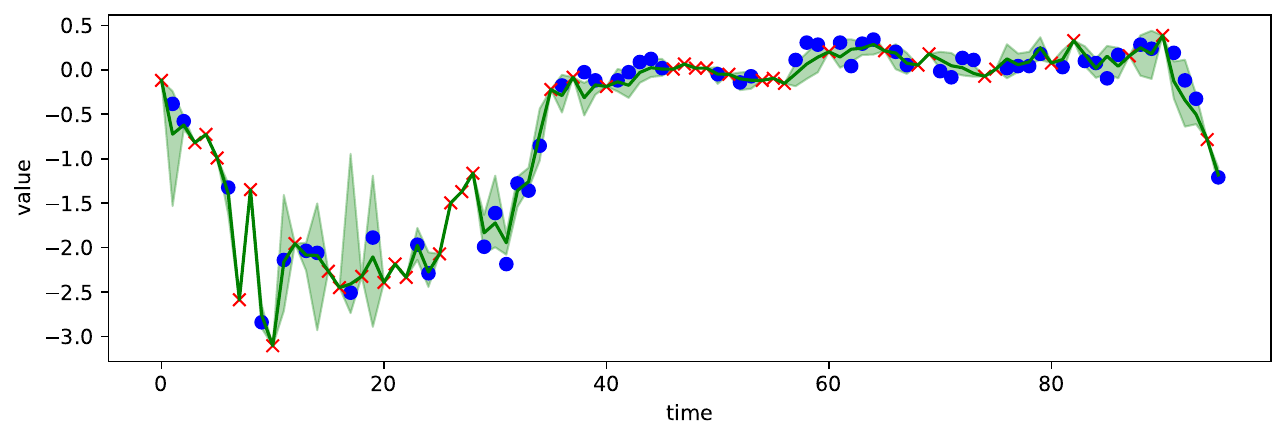}
\end{minipage}%
}%

\subfigure[Imputation results on ETTm2 dataset.]{
\begin{minipage}{\linewidth}
\centering
\includegraphics[width=5in]{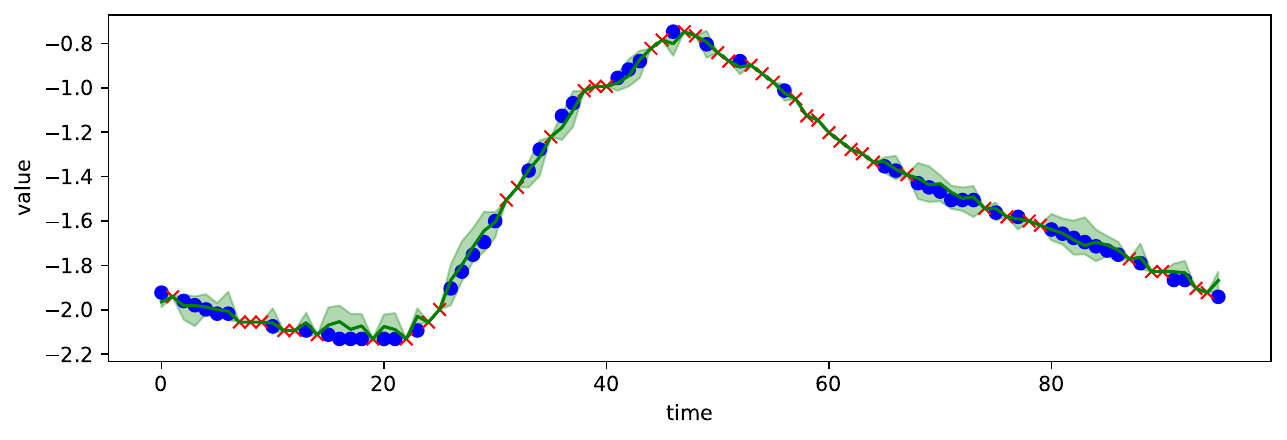}
\end{minipage}%
}%

\caption{Visualization of imputation task on ETT datasets. This figure illustrates TimeDiT's performance, with red $\times$'s marking observed values, blue dots showing ground truth points for interpolation, a green line representing TimeDiT's mean of interpolation, and green shading indicating its estimated uncertainty intervals. }
\label{impu_fig1}
\end{figure}

\begin{figure}[htbp]
\centering
\subfigure[Imputation results on electricity dataset]{
\begin{minipage}{\linewidth}
\centering
\includegraphics[width=5in]{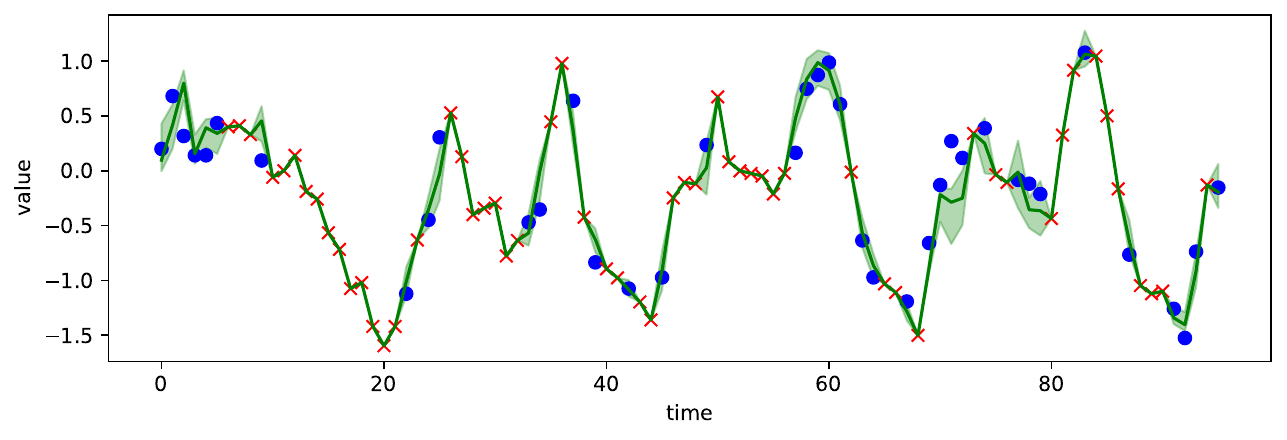}
\end{minipage}%
}

\subfigure[Imputation results on weather dataset]{
\begin{minipage}{\linewidth}
\centering
\includegraphics[width=5in]{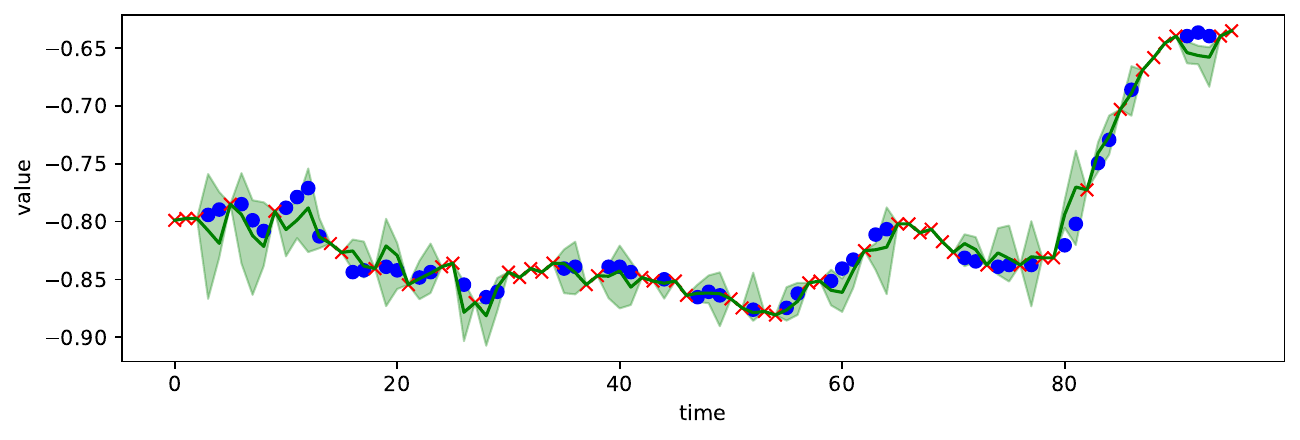}
\end{minipage}%
}

\caption{Visualization of imputation task on electricity and weather datasets. This figure illustrates TimeDiT's performance, with red $\times$'s marking observed values, blue dots showing ground truth points for interpolation, a green line representing TimeDiT's mean of interpolation, and green shading indicating its estimated uncertainty intervals. }
\label{impu_fig2}
\end{figure}

\subsection{Imputation}
\label{impute_details}

\subsubsection{Full Imputation Results}
The imputation task results, presented in Table~\ref{tab:full_impute}, demonstrate TimeDiT's superior performance across various datasets and missing data ratios. All baseline models are trained in a full-shot setting, while TimeDiT leverages a pre-trained
foundation model, fine-tuning it on realistic datasets. TimeDiT consistently achieves the lowest Mean Squared Error (MSE) and Mean Absolute Error (MAE) scores in most scenarios, outperforming state-of-the-art models such as GPT2, TimesNet, and PatchTST. Notably, TimeDiT's performance remains robust even as the proportion of missing data increases from 12.5\% to 50\%, showcasing its ability to handle substantial data gaps effectively. The model's imputation accuracy is particularly impressive for the ETTh1, ETTh2, ETTm1, and ETTm2 datasets, where it maintains a significant lead over other methods.  {imeDiT demonstrates superior performance on most datasets, achieving significant improvements over Timer, TimeMixer, and iTransformer, particularly on ETT datasets where we see reductions in MSE by up to 60\%. TimeDiT maintains strong overall performance while offering greater versatility}

\subsubsection{Imputation Visualization}
For visual representation of TimeDiT's imputation capabilities, we have plotted the results in Figure ~\ref{impu_fig1} and Figure~\ref{impu_fig2}, which clearly illustrates the model's accuracy in reconstructing missing data points across different datasets and missing data ratios.

\begin{figure}[t]
    \centering
    \includegraphics[width = \linewidth]{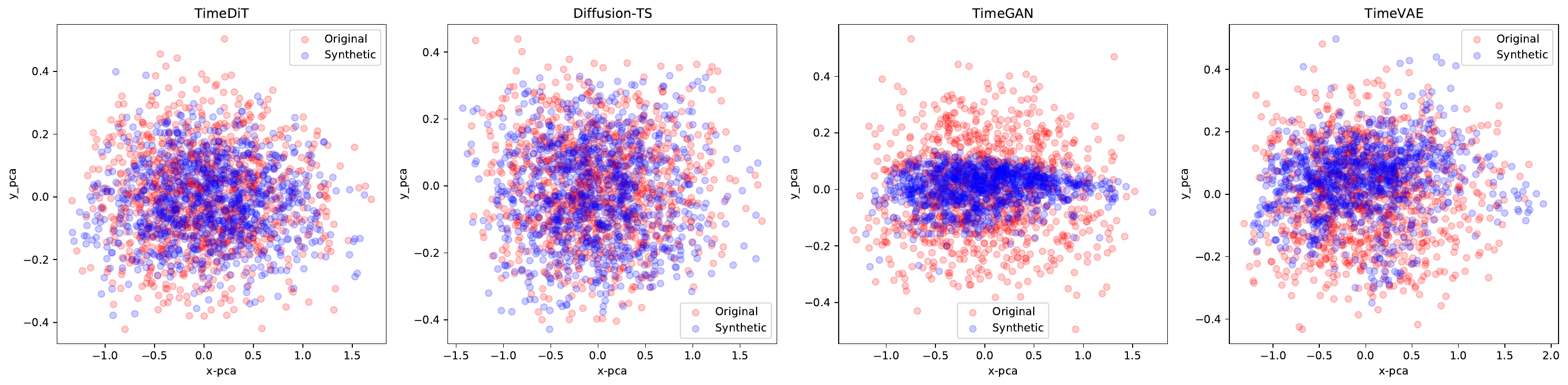}
    \caption{PCA Evaluation of Synthetic TSD from TimeDiT and other baselines on the sine dataset.}
    \label{fig:sine-pca}
\end{figure}
\begin{figure}[t]
    \centering
    \includegraphics[width = \linewidth]{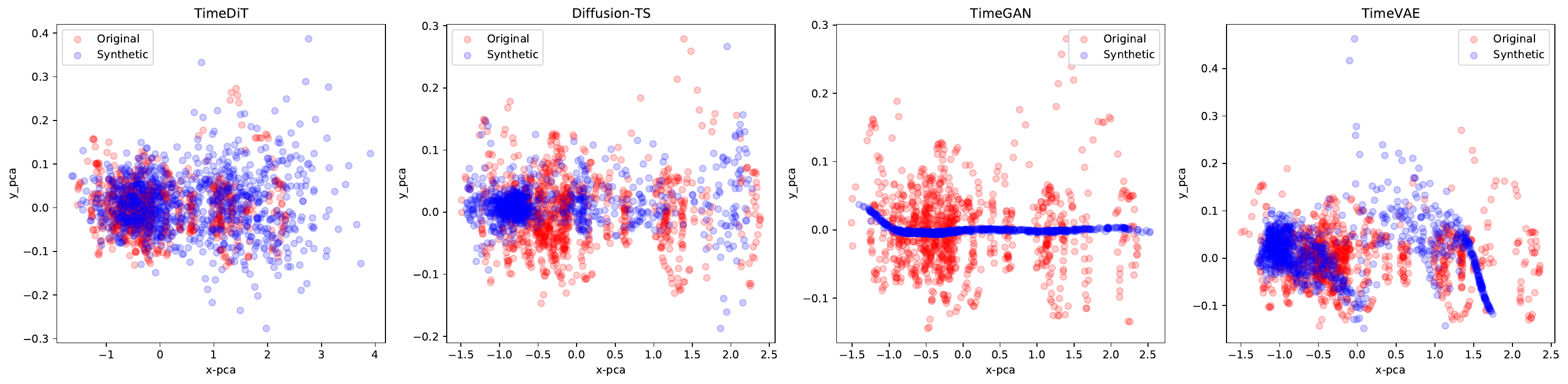}
    \caption{PCA Evaluation of Synthetic TSD from TimeDiT and other baselines on the stock dataset.}
    \label{fig:stock-pca}
\end{figure}
\begin{figure}[t]
    \centering
    \includegraphics[width = \linewidth]{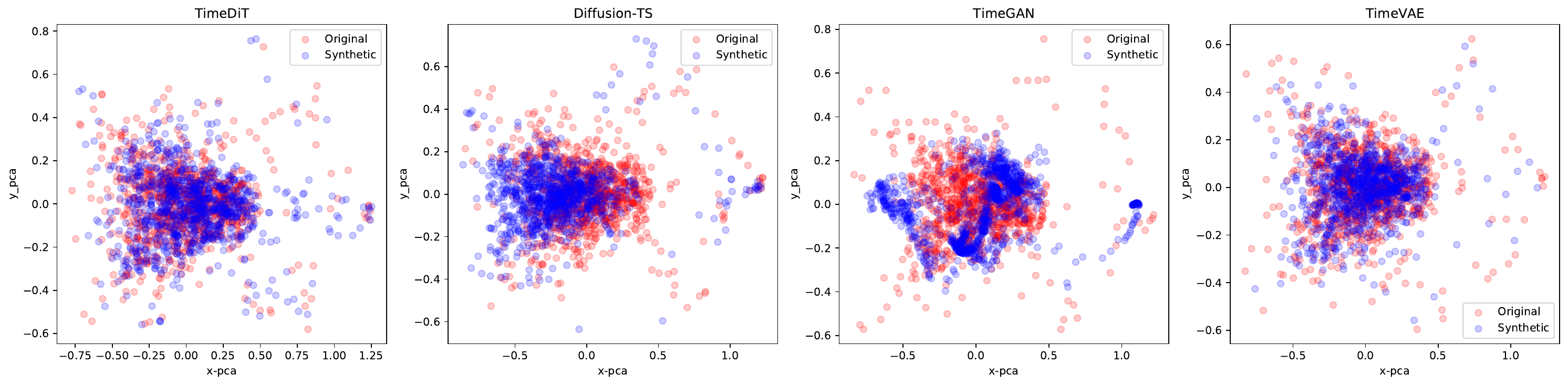}
    \caption{PCA plot for air quality dataset. }
    \label{fig:air-pca}
\end{figure}
\begin{figure}[t]
    \centering
    \includegraphics[width = \linewidth]{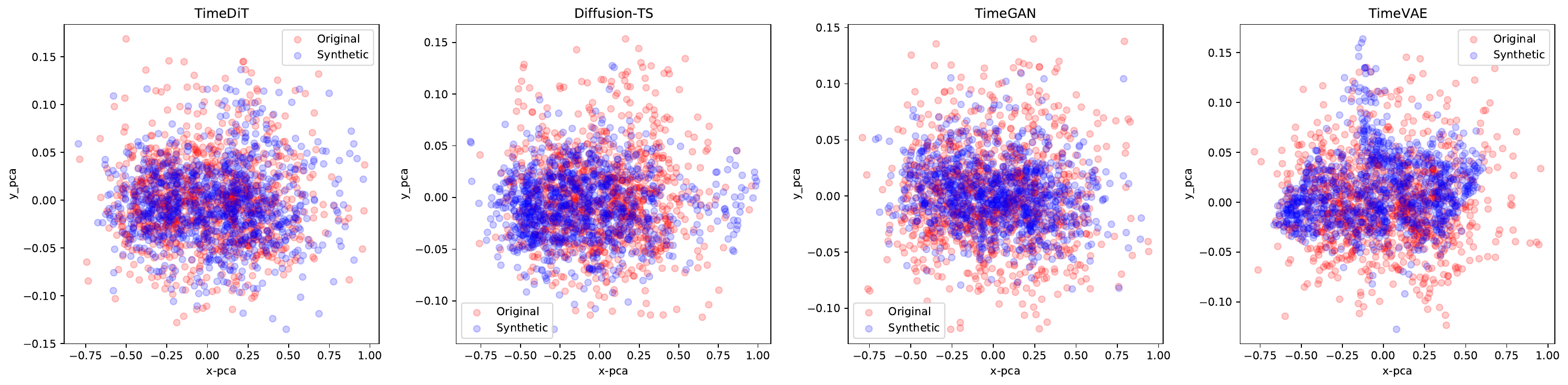}
    \caption{PCA plot for energy dataset. }
    \label{fig:energy-pca}
\end{figure}

\subsection{Synthetic Generation} \label{synthetic_visual}

\subsubsection{Synthetic Generation Visualization}
We use 80\% of all data for training and evaluation of the same data. For the air quality dataset, previous methods did not carefully use the -200 values as a placeholder for missing values. In our experiment, we masked all the -200 values for \modelnamespace and baselines that support masks. For baselines that do not support mask, we replace -200 with the mean value. minmax scaler is used for all models. Diffusion-TS uses a different normalization scheme between -1 and 1. We replace its normalization scheme to be minmax scaler to ensure fair comparison. Figure \ref{fig:sine-pca}, \ref{fig:stock-pca},\ref{fig:air-pca},\ref{fig:energy-pca} shows the PCA plots for all datasets and baselines. The visual comparison also validates the superiority of \modelname. 

\subsubsection{Limited Synthetic Generation}
We also run the generation experiments with the limited data fine-tuning in Table~\ref{tab:syn_2}.
The generation experiments with limited data fine-tuning demonstrate TimeDiT's superior performance across various datasets and evaluation metrics. Comparing TimeGAN, TimeVAE, Diffusion-TS, and TimeDiT on sine, air, and energy datasets with 5\% and 10\% training data, TimeDiT consistently achieves the lowest Discriminative Scores, indicating its ability to generate the most realistic time series. In terms of Predictive Scores, TimeDiT outperforms or matches other models, particularly excelling in the air dataset. Notably, TimeDiT's performance remains robust or improves when increasing from 5\% to 10\% training data, showcasing its effectiveness in data-scarce scenarios. These results highlight TimeDiT's capability to capture complex temporal patterns and generate high-quality time series data, even with limited training samples, making it a promising tool for various time series generation tasks.

\begin{table}[t]
\centering
\caption{Limited observation data Synthetic Generation results on 24-length multivariate time series. Discriminative and predictive scores are calculated as described in \citep{yoon2019time}.}
\scalebox{0.8}{
\begin{tabular}{cccccccc}
\toprule
\multirow{2}{*}{\textbf{Metric}} & \multirow{2}{*}{\textbf{Methods}} &\multicolumn{3}{c}{\textbf{0.05}} & \multicolumn{3}{c}{\textbf{0.1}} \\\cline{3-8}
& & \textbf{Sine} & \textbf{Air Quality} & \textbf{Energy}  & \textbf{Sine} & \textbf{Air Quality} & \textbf{Energy} \\ 

\midrule
\multirow{4}{*}{\shortstack{Discriminative \\ Score}} 
& TimeGAN & 0.120(0.043) & 0.500(0.003) & 0.500(0.000) & 0.067(0.028) & 0.492(0.003) & 0.500(0.000) \\
& TimeVAE & 0.220(0.224) & 0.498(0.001) & 0.500(0.000) & 0.499(0.002) & 0.495(0.002) & 0.499(0.001) \\
& Diffusion-TS & 0.037(0.013) & 0.496(0.003) & 0.498(0.005) & 0.031(0.012) & 0.494(0.001) & 0.494(0.011) \\
& TimeDiT & \textbf{0.031(0.007)} & \textbf{0.456(0.003)} & \textbf{0.472(0.000)} & \textbf{0.030(0.009)} & \textbf{0.437(0.004)} & \textbf{0.447(0.002)} \\
\midrule
\multirow{4}{*}{\shortstack{Predictive \\ Score}} 
& TimeGAN & 0.231(0.007) & 0.148(0.029) & 0.308(0.006) & 0.200(0.002) & 0.130(0.029) & 0.302(0.004) \\
& TimeVAE & 0.251(0.003) & 0.328(0.008) & \textbf{0.296(0.001)} & 0.238(0.002) & 0.308(0.014) & \textbf{0.288(0.001)} \\
& Diffusion-TS & 0.196(0.003) & 0.111(0.004) & 0.333(0.018) & 0.188(0.001) & 0.102(0.010) & 0.340(0.019) \\
& TimeDiT & \textbf{0.194(0.001)} & \textbf{0.089(0.005)} & 0.335(0.008) & \textbf{0.192(0.000)} &\textbf{ 0.070(0.007)} & 0.318(0.005) \\
\bottomrule
\end{tabular}}
\label{tab:syn_2}
\end{table}

\subsection{Anomaly Detection} \label{anomaly_details}
We conduct experiments on five real-world datasets from industrial applications: MSL, SMAP, SWaT, SMD, and PSM.  The diffusion model, renowned for its proficiency in distribution learning, may inadvertently overfit by reconstructing anomalies alongside normal data points. To counteract this, we opted to bypass pretraining and introduced spectral residue (SR) transformation at the preprocessing stage of \modelname. This transformation helps to conceal points most likely to be anomalies and their immediate neighbors. The number of neighbors affected is controlled by the hyperparameter $n_{neighbor}$. The SR method utilizes Fourier Transformation to convert the original time series into a saliency map, thereby amplifying abnormal points, as detailed in \citep{ren2019time, zhao2020multivariate}.
Consistent with prior methodologies, we set the sequence length to be 100 identify anomalies using the 99th percentile of reconstruction errors. During evaluations, we apply standard anomaly adjustments as suggested by \citep{xu2018unsupervised}. As demonstrated in Table \ref{tab:ad_main}, \modelnamespace outperforms baseline models on four of the five datasets. In particular, \modelnamespace 23.03 points of improvement in terms of F1 score on the SMAP dataset compared to the previous best baseline. { In addition, TimeDiT consistently outperforms both TimeMixer and iTransformer across all datasets, with particularly notable improvements on SMAP (95.91 vs 67.63/66.76) and SWAT (97.57 vs 88.84/92.63). These comprehensive comparisons against the latest models demonstrate TimeDiT's effectiveness as a unified framework for time series analysis, often achieving state-of-the-art performance while maintaining broader applicability across diverse tasks.}

\begin{table}[b]
\centering

\label{tab:full_ad}
\caption{Anomaly Detection result on 100-length multivariate time series. We calculate Precision, Recall, and F1 score as \% for each dataset. '$.$' notation in model name stands for transformer. \textbf{Bold} indicates best result, \underline{Underline} indicates the second best result. We replace the joint criterion in Anomaly Transformer with reconstruction error for fair comparison.}

\resizebox{\textwidth}{!}{
\begin{tabular}{lcccccccccccccccc}
\toprule
\textbf{Methods}& \multicolumn{3}{c}{\textbf{MSL}} & \multicolumn{3}{c}{\textbf{SMAP}} & \multicolumn{3}{c}{\textbf{SWaT}}  & \multicolumn{3}{c}{\textbf{SMD}} & \multicolumn{3}{c}{\textbf{PSM}}  & \textbf{1st Pl} \\
\textbf{Metrics} & \textbf{P} & \textbf{R} & \textbf{F1} & \textbf{P} & \textbf{R} & \textbf{F1} & \textbf{P} & \textbf{R} & \textbf{F1} & \textbf{P} & \textbf{R} & \textbf{F1} & \textbf{P} & \textbf{R} & \textbf{F1}  &\textbf{Count} \\
\midrule
\modelname & \textbf{91.54} & \underline{87.23} & \textbf{89.33} & \textbf{93.35} & \textbf{98.61} & \textbf{95.91} & \textbf{93.64} & \textbf{99.46} & \textbf{96.46} & 78.83 & \textbf{88.26} & 83.28 & 97.36 & \textbf{97.79} & \textbf{97.57} & {11}\\
\midrule
GPT(6) & 82.00 & 82.91 & 82.45 & 90.60 & \underline{60.95} & \underline{72.88} & \underline{92.20} & 96.34 & 94.23 & \underline{88.89} & \underline{84.98} & \textbf{86.89} & 98.62 & 95.68 & 97.13  & 1\\
TimesNet & \underline{89.54} & 75.36 & 81.84 & 90.14 & 56.40 & 69.39 & 90.75 & 95.40 & 93.02 & 87.91 & 81.54 & 84.61 & 98.51 & 96.20 & \underline{97.34} &0\\
PatchTST & 88.34 & 70.96 & 78.70 & 90.64 & 55.46 & 68.82 & 91.10 & 80.94 & 85.72 & 87.26 & 82.14 & 84.62 & 98.84 & 93.47 & 96.08 &0\\
ETSformer & 85.13 & 84.93 & \underline{85.03} & 92.25 & 55.75 & 69.50 & 90.02 & 80.36 & 84.91 & 87.44 & 79.23 & 83.13 & \textbf{99.31} & 85.28 & 91.76 & 1 \\
FEDformer & 77.14 & 80.07 & 78.57 & 90.47 & 58.10 & 70.76 & 90.17 & 96.42 & 93.19 & 87.95 & {82.39} & 85.08 & 97.31 & \underline{97.16} & 97.23 &0 \\
LightTS & 82.40 & 75.78 & 78.95 & \underline{92.58} & 55.27 & 69.21 & 91.98 & 94.72 & 93.33 & 87.10 & 78.42 & 82.53 & 98.37 & 95.97 & 97.15 &0 \\
DLinear & 84.34 & 85.42 & 84.88 & 92.32 & 55.41 & 69.26 & 80.91 & 95.30 & 87.52 & 83.62 & 71.52 & 77.10 & 98.28 & 89.26 & 93.55 &0\\
Autoformer & 77.27 & 80.92 & 79.05 & 90.40 & 58.62 & 71.12 & 89.85 & 95.81 & 92.74 & 88.06 & 82.35 & 85.11 & \underline{99.08} & 88.15 & 93.29 &0 \\
Anomaly. & 79.61 & \textbf{87.37} & 83.31 & 91.85 & 58.11 & 71.18 & 72.51 & \underline{97.32} & 83.10& \textbf{88.91} & 82.23 & \underline{85.49}  & 68.35 & 94.72 & 79.40&2 \\
{TimeMixer}&{89.72} &{75.42} &{81.95} &{89.51} &{54.34} &{67.63} &{91.56} &{86.28} &{88.84} &{86.60} &{71.50} &{78.33} &{99.18} &{87.74} &{93.11} & {0}
 \\
{iTransformer}&{86.16} &{62.64} &{72.54} &{90.69} &{52.82} &{66.76} &{92.21} &{93.06} &{92.63} &{86.92} &{77.75} &{82.08} &{97.98} &{92.81} &{95.32} & {0}
\\

\bottomrule

\end{tabular}}
\end{table}

\begin{table}[ht]
\centering
\caption{{Threshold Sensitivity Analysis on Anomaly Detection Performance evaluated on F1 score}}
\label{App:ad_thre}
\begin{tabular}{lcccccc}
\toprule
Threshold & 99.5 & 99 & 98 & 97 & 96 & 95 \\
\hline
MSL & 83.9 & 89.33 & \textbf{90.1} & 88.17 & 85.28 & 82.84 \\
PSM & 96.32 & \textbf{97.57} & 96.78 & 95.72 & 94.66 & 93.61 \\
SMAP & \textbf{97.08} & 95.91 & 93.23 & 90.33 & 87.64 & 85.09 \\
SMD & \textbf{83.28} & 82.07 & 76.61 & 70.73 & 65.71 & 61.24 \\
SWAT & \textbf{97.6} & 96.46 & 93.49 & 90.74 & 88.0 & 85.42 \\
\bottomrule
\end{tabular}
\end{table}

{\paragraph{Anomaly Detection Threshold} Our comprehensive analysis of threshold selection in Table~\ref{App:ad_thre} revealed that higher percentile thresholds, particularly the 99th and 99.5th percentiles, consistently yield superior performance. While we observed a systematic degradation in detection accuracy as threshold values decrease, we maintained the 99th percentile threshold to ensure fair comparison with existing methodologies. This decision reflects our commitment to methodological rigor, as optimizing threshold values based on test set performance would introduce bias in the comparative analysis. Our approach prioritizes consistent experimental conditions across all evaluated methods, enabling meaningful benchmark comparisons while acknowledging the impact of threshold selection on detection performance.}

\paragraph{Spectral Residue processing for Anomaly Detection. } The SR Transformation involves the following equations. Table \ref{tab:full_ad} shows the full anomaly detection results. 
\begin{equation}
A(f) = \text{Amplitude}(F(x))
\end{equation}
\begin{equation}
P(f) = \text{Phase}(F(x))
\end{equation}
\begin{equation}
L(f) = \log(A(f))
\end{equation}
\begin{equation}
AL(f) = h_q(f) \cdot L(f)
\end{equation}
\begin{equation}
R(f) = L(f) - AL(f)
\end{equation}
\begin{equation}
S(x) = F^{-1}(\exp(R(f) + iP(f)))
\end{equation}

\section{Analysis on TimeDiT}

We present a comprehensive analysis of TimeDiT's design space, conducting systematic comparisons across different architectural variants. To ensure fair evaluation, all experiments maintain consistent training configurations, utilizing the same checkpoint and number of training steps. This rigorous experimental setup allows us to isolate and assess the impact of individual architectural components while controlling for training conditions.

\subsection{{Ablation Study}}

\begin{table}[ht]
\caption{{Results on the Model Design Space, fair comparison on the same checkpoint with same training steps.}}
\centering
\begin{tabular}{lccccc}
\hline
Dataset & TimeDiT &  Dual-attention & Channel-wise & Patch Token \\
\hline
Solar & 0.457(0.002) & 0.467(0.002) & 0.461(0.003) & 0.874(0.010) \\
Electricity & 0.026(0.001) & 0.029(0.001) & 0.028(0.000) & 0.105(0.013) \\
Traffic & 0.185(0.010) &  0.187(0.007) & 0.164(0.006) & 0.258(0.021) \\
\hline
\end{tabular}
\end{table}

{Our comprehensive ablation studies, detailed in Sections E1, E2, and E3, systematically evaluate TimeDiT's architectural choices. In Section E1, with particular emphasis on the Transformer design strategy, we explore TimeDiT's temporal-wise attention mechanism and compare it against alternative approaches, including channel-wise attention and dual attention mechanisms (as discussed in ~\citep{PMLR-v235-yu24s}). The analysis demonstrates that temporal-wise processing significantly outperforms traditional patch-based tokenization approaches, achieving substantially lower error rates (0.457 versus 0.874 on Solar dataset).}

{This performance disparity can be attributed to two key factors: First, while channel relationships exhibit model-specific variations, temporal patterns provide more universal characteristics across time series data, enabling better generalization. Second, patch-based approaches introduce additional hyperparameter dependencies (patch length and stride settings) that compromise the model's universal applicability. These findings validate our design choice of temporal-wise processing as a more robust and generalizable approach for time series modeling. The empirical results strongly support our architectural decisions, demonstrating that TimeDiT's temporal-focused design effectively captures universal temporal dynamics while maintaining model flexibility across diverse applications and domains. In addition, the Physics-Informed component yields consistent performance improvements across all datasets, with notable enhancements in Traffic (0.153 versus 0.185), Electricity (0.024 versus 0.026), and Solar (0.452 versus 0.457) predictions, underscoring the value of incorporating physical constraints during inference.}

\subsection{{Handling missing values}}

\begin{table}[ht]
\centering
\caption{{Mask mechanisms for TimeDiT, compared on the zero-shot forecasting task, , fair comparison on the same checkpoint with same training steps.}}
\label{App_masks}
\begin{tabular}{lcccc}
\hline
Dataset & TimeDiT & w/o Random Mask & w/o Stride Mask & w/o Block Mask \\
\hline
Solar & \textbf{0.457(0.002)} & {0.463(0.002)} & 0.465(0.002) & 0.843(0.005) \\
Electricity & \textbf{0.026(0.001)} & 0.029(0.001) & 0.030(0.001) & 0.095(0.006) \\
Traffic & \textbf{0.185(0.010)} & 0.191(0.007) & 0.188(0.007) & 0.201(0.011) \\
\hline
\end{tabular}
\end{table}

{Our experimental design leverages naturally occurring missing values inherent in real-world datasets, primarily arising from irregular sampling rates and multi-resolution data collection processes. This approach authentically validates model robustness against genuine missing data patterns rather than artificially generated scenarios. TimeDiT incorporates a comprehensive masking strategy that aligns with three well-established missing data mechanisms: Missing Completely at Random (MCAR) using uniform distribution-based random masks, Missing at Random (MAR) employing block and stride masks to capture structured patterns and dependencies between non-contiguous observations, and Missing Not at Random (MNAR) utilizing reconstruction masks with physics-informed sampling for scenarios where missing patterns correlate with unobserved variables. These mechanisms are simultaneously applied through self-supervised learning, enabling robust representation learning without requiring explicit knowledge of the underlying missing data processes. Our comprehensive ablation studies in Table~\ref{App_masks} demonstrate the criticality of each masking strategy, where removing any mask type leads to performance degradation, with future masks showing the most significant impact. These findings validate our integrated approach to handling diverse missing data scenarios in time-series analysis.}

\subsection{{Condition scheme for TimeDiT}}

\begin{table}[ht]
\centering
\label{tab:condit}
\caption{{Condition scheme for TimeDiT, compared on the zero-shot forecasting task, fair comparison on the same checkpoint with same training steps.}}
\begin{tabular}{lcccc}
\hline
Dataset & AdaLN & Additive & Cross-attention & Token concatenation \\
\hline
Solar & \textbf{0.457(0.002)} & 0.671(0.002) & 0.721(0.002) & 0.463(0.001) \\
Electricity & \textbf{0.026(0.001)} & 0.068(0.004) & 0.079(0.003) & 0.041(0.003) \\
Traffic & \textbf{0.185(0.010)} & 0.224(0.001) & 0.216(0.000) & 0.188(0.008) \\
\hline
\end{tabular}
\end{table}

{As mentioned in Section~\ref{sec_diff_trans}, AdaLN's superior performance stems from its ability to dynamically adjust feature distributions across different layers while maintaining computational efficiency. This approach aligns well with the inherent nature of time series data, where temporal dependencies typically exhibit gradual rather than dramatic changes in both seen and unseen time steps. We conducted comparative experiments to evaluate different conditioning mechanisms in TimeDiT:
\begin{itemize}
    \item Additive conditioning, which adds conditional information directly to the diffusion input; 
    \item Cross-attention, which uses conditional time series as keys/values and noisy time series as queries to fuse conditional information;
    \item Token concatenation, which concatenates conditional time series with noisy time series at the input level before TimeDiT processing. 
\end{itemize}
The experimental results (Table~\ref{tab:condit}) across Solar, Electricity, and Traffic datasets consistently show that AdaLN achieves superior performance compared to the next best alternative. This significant performance gap validates our choice of AdaLN as TimeDiT's primary conditioning mechanism.}

\subsection{{Noise Embedding Justification}}
{TimeDiT's noise embedding approach plays multiple key roles in the diffusion modeling framework. The diffusion process operates directly in a continuous embedding space, allowing for smoother transitions between noise levels and better preserving the inherent time dependence, thus enabling the model to learn a more robust representation of the underlying time series structure. This approach has several technical advantages~\citep{ho2020denoising, Peebles2022DiT, lu2024vdt}: the embedding space provides a continuous representation in which the diffusion process can operate more efficiently. 
The direct embedding of noisy samples helps prevent the embedding space from collapsing during training. From a practical point of view, this approach allows for parallel processing of multiple time steps, handles varying degrees of noise through a unified framework, and makes the diffusion process more stable compared to traditional generation methods. In addition, the embedded noise representation allows for the seamless incorporation of physical constraints and maintains temporal continuity while progressively denoising, thus contributing to a better quantification of the uncertainty in the generated samples. Direct prediction of the input is also an option available, and we added new experiments as shown in Table~\ref{app_direct_input}. This also demonstrates the advantages of reconstructive noise.}
\begin{table}
\centering
\caption{{Results in predicting the input of TimeDiT, compared on the zero-shot forecasting task,  fair comparison on the same checkpoint with same training steps.}}
\label{app_direct_input}
\begin{tabular}{lcc}
\toprule
Dataset & TimeDiT & Predict the input \\
\hline
Solar & 0.457(0.002) & 0.462(0.003) \\
Electricity & 0.026(0.001) & 0.037(0.002) \\
Traffic & 0.185(0.010) & 0.199(0.007) \\
\bottomrule
\end{tabular}
\end{table}

\subsection{{Failure  Scenarios Analysis}}

{TimeDiT's performance shows notable degradation in three key scenarios: highly irregular sampling rates deviating from training distributions, complex non-stationary patterns underrepresented in pretraining data, and domain-specific patterns requiring expert knowledge beyond general time series characteristics. As shown by SMD dataset for anomaly detection (Table ~\ref{tab:ad_main}) where it achieves 83.28\% F1 score versus GPT2's 86.89\%. This dataset represents cloud server machine metrics with high-frequency sampling and complex feature interdependencies. Additionally, when dealing with extremely short-term patterns or highly localized anomalies, specialized architectures like GPT2 that focus intensively on recent temporal context may outperform TimeDiT's more holistic approach, as our diffusion-based generation process may occasionally smooth over abrupt local changes. These limitations, primarily stemming from the model's dependence on learned foundational patterns, become particularly relevant in specialized industrial applications and unique financial scenarios. Understanding these boundaries is crucial for informed model deployment decisions and highlights promising directions for future research.}

\subsection{Dynamic on Model Size}

The experimental results demonstrate a clear correlation between TimeDiT's model size and its imputation performance across different datasets and missing data ratios. As shown in Table~\ref{label_size}, as the model size increases from Small (S) to Big (B) to Large (L), we observe consistent improvements in both averaged  Mean Squared Error (MSE) and averaged Mean Absolute Error (MAE) metrics. The Large model consistently outperforms the Small and Big variants across all scenarios, with the most significant gains observed in the weather dataset. Notably, larger models (B and L) show better resilience to increased proportions of missing data compared to the Small model. The improvement is more pronounced for the weather dataset than for the ecl dataset, suggesting that the benefits of increased model size may vary depending on the nature and complexity of the time series data. The consistent performance gains from S to B to L models indicate that TimeDiT's architecture scales well with increased model size. These findings suggest that increasing TimeDiT's model size is an effective strategy for improving imputation accuracy, particularly for complex datasets or scenarios with higher proportions of missing data. However, the performance may remain relatively consistent across all model sizes for both the weather and ecl datasets, even as the proportion of missing data increases from 12.5\% to 50\%. This stability in performance suggests that TimeDiT's architecture may achieve its optimal capacity for these imputation tasks even at smaller model sizes.  Thus, the trade-off between computational resources and performance gains should be considered when selecting the appropriate model size for specific applications.

\begin{table}[t]
\centering
\caption{Performance metrics for weather and ecl datasets on different model size.}
\scalebox{0.9}{
\begin{tabular}{cccccccc}
\toprule
 & & \multicolumn{2}{c}{\textbf{S}} & \multicolumn{2}{c}{\textbf{B}} & \multicolumn{2}{c}{\textbf{L}} \\
\cmidrule(lr){3-4} \cmidrule(lr){5-6} \cmidrule(lr){7-8}
& & \textbf{MSE} & \textbf{MAE} & \textbf{MSE} & \textbf{MAE} & \textbf{MSE} & \textbf{MAE} \\
\midrule
\multirow{5}{*}{\textbf{Weather}} & 0.125 & 0.029 & 0.033 & 0.029 & 0.026 & 0.025 & 0.024 \\
 & 0.250 & 0.031 & 0.033 & 0.033 & 0.029 & 0.028 & 0.027 \\
 & 0.375 & 0.034 & 0.037 & 0.036 & 0.033 & 0.031 & 0.031 \\
 & 0.500 & 0.031 & 0.041 & 0.042 & 0.039 & 0.036 & 0.036 \\
 & \textbf{Avg} & 0.031 & 0.036 & 0.035 & 0.032 & 0.030 & 0.029 \\
\midrule
\multirow{5}{*}{\textbf{ECL}} & 0.125 & 0.051 & 0.148 & 0.050 & 0.144 & 0.048 & 0.140 \\
 & 0.250 & 0.061 & 0.163 & 0.060 & 0.158 & 0.058 & 0.154 \\
 & 0.375 & 0.074 & 0.181 & 0.071 & 0.175 & 0.069 & 0.170 \\
 & 0.500 & 0.090 & 0.202 & 0.087 & 0.197 & 0.084 & 0.190 \\
 & \textbf{Avg} & 0.069 & 0.174 & 0.067 & 0.169 & 0.065 & 0.163 \\
\bottomrule
\end{tabular}}
\label{label_size}
\end{table}

\subsection{Learned Representation}
\begin{figure*}[htbp]
\centering
\subfigure[TimeDiT without textual information]{
\begin{minipage}{0.5\linewidth}
\centering
\includegraphics[width=\textwidth]{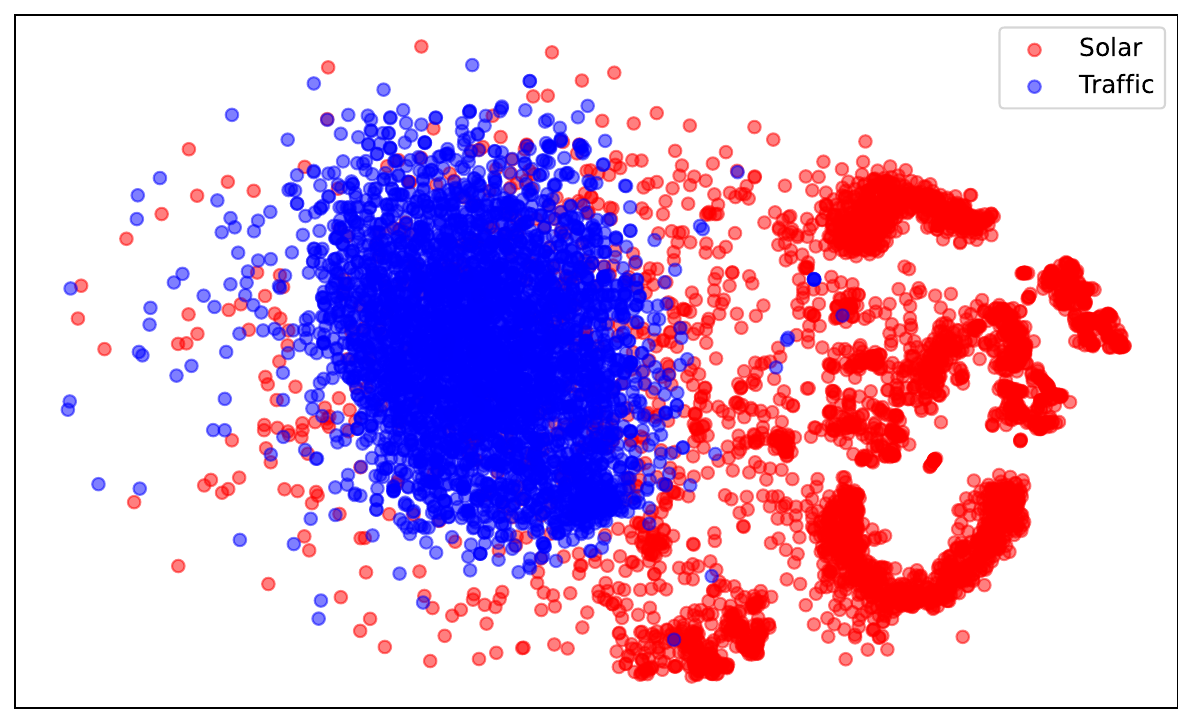}
\end{minipage}%
}%

\subfigure[TimeDiT with textual information]{
\begin{minipage}{0.5\linewidth}
\centering
\includegraphics[width=\textwidth]{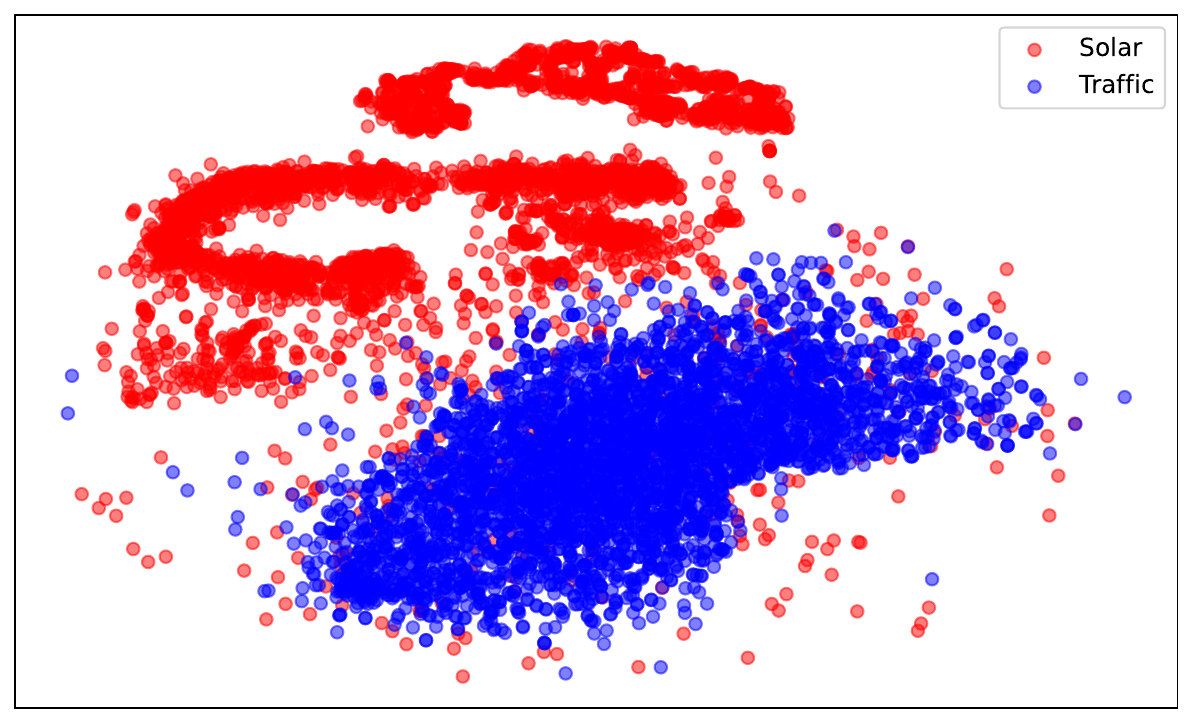}
\end{minipage}%
}%
\caption{Repreasentation Visualization of TimeDiT when the input is two new  datasets and it can modeling them separately with the capability. }
\label{rep_fig}
\end{figure*}

We randomly sampled 4000 training samples from each of the Solar and Traffic datasets and got their representation from the foundation model with and without textual condition, which is the zero-shot setting. To visualize the distribution of these datasets, we employ t-SNE dimensionality reduction. As depicted in Figure~\ref{rep_fig}, the t-SNE plot clearly distinguishes between the Solar and Traffic datasets, highlighting their unique characteristics. The Solar dataset samples form a distinct cluster, likely reflecting the periodic patterns and seasonal variations inherent in solar power generation. In contrast, the Traffic dataset samples create a separate cluster, capturing the complex temporal dynamics of traffic flow, which may include daily commute patterns and irregular events. This clear separation in the t-SNE visualization underscores the fundamental differences in the underlying structures and patterns of these two time series datasets. Such distinction is crucial for understanding the diverse nature of temporal data and highlights the importance of developing versatile models like TimeDiT that can effectively capture and generate a wide range of time series patterns.

\end{document}